\documentclass[num-refs]{nbdt-article}
\pdfoutput=1

\usepackage{siunitx}
\usepackage[utf8]{inputenc}

\papertype{Original Article}
\paperfield{Journal Section}

\title{Estimating smooth and sparse neural receptive fields with a flexible spline basis}

\author[1]{Ziwei Huang}
\author[1,2]{Yanli Ran}
\author[1]{Jonathan Oesterle}
\author[1,2]{Thomas Euler}
\author[1-3]{Philipp Berens}

\affil[1]{Institute for Ophthalmic Research, University of Tübingen, Tübingen, Germany}
\affil[2]{Centre for Integrative Neuroscience, University of Tübingen, Tübingen, Germany}
\affil[3]{Tübingen AI Center, University of Tübingen, Tübingen, Germany}

\corraddress{Philipp Berens, Institute for Ophthalmic Research, University of Tübingen, Tübingen, Germany}
\corremail{philipp.berens@uni-tuebingen.de}

\fundinginfo{The German Research Foundation: BE5601/4-1 and BE5601/8-1; the Collaborative Research Center 1233 ``Robust Vision'': 276693517; Individual research grants: EU 42/10-1 and BE5601/6-1; the German Ministry of Education and Research: 01GQ1601.}

\runningauthor{Huang et al.}

\abbrevs{ALD, automatic locality determination; ASD, automatic smoothness determination; LG, Linear-Gaussian model; LNLN, Linear-Nonlinear cascade model; LNP, Linear-Nonlinear-Poisson model;  MAP, maximum a posteriori; MSE, mean squared error; MLE, maximum-likelihood estimate; NMF, nonnegative matrix factorization; RGC, retinal ganglion cell; STA, spike-triggered average; STRF, spatio-temporal receptive field; wSTA, whitened-STA.}

\begin{document}

\maketitle
\begin{abstract}

Spatio-temporal receptive field (STRF) models are frequently used to approximate the computation implemented by a sensory neuron. Typically, such STRFs are assumed to be smooth and sparse. Current state-of-the-art approaches for estimating STRFs based on empirical Bayes are often not computationally efficient in high-dimensional settings, as encountered in sensory neuroscience. Here we pursued an alternative approach and encode prior knowledge for estimation of STRFs by choosing a set of basis functions with the desired properties: natural cubic splines. Our method is computationally efficient and can be easily applied to a wide range of existing models. We compared the performance of spline-based methods to non-spline ones on simulated and experimental data, showing that spline-based methods consistently outperform the non-spline versions.

\end{abstract}

\section{Introduction}

Spatio-temporal receptive fields (STRFs) are frequently used in neuroscience to approximate the computation implemented by a sensory neuron. Such models consist typically of one or more linear filters summing up sensory inputs across time and space, followed by a static nonlinearity and a probabilistic output process \cite{aljadeff_analysis_2016}. For this, different distributions can be used depending on the type of recorded neuronal responses.

The simplest way to estimate a STRF for a given data set is arguably to compute the spike-triggered average (STA), the average over all stimuli preceding a spike \cite{schwartz_spike-triggered_2006}. However, the results of the spike-triggered analysis are often noisy due to limited data and prone to be distorted when the neuron is stimulated with correlated noise or natural stimuli \cite{paninski_convergence_2003}. The maximum likelihood estimate (MLE), also known as whitened-STA (wSTA) \cite{theunissen_estimating_2001}, does not suffer from such shortcomings, but it requires even more data to converge as the whitening procedure tends to amplify noise at frequencies with low power, making it a less preferred option in many experimental settings. A modified wSTA that removes low power frequencies can result in smoother STRF, while the threshold for such a frequency cutoff is normally chosen by cross-validation \cite{theunissen_estimating_2001}.

It has been suggested that these shortcomings can be overcome by computing the maximum a posteriori (MAP) estimates within the framework of Bayesian inference \cite{wu_complete_2006}. Under this framework, prior knowledge about STRFs such as sparsity, smoothness \cite{sahani_evidence_2002}, and locality\cite{park_receptive_2011} can be encoded into a prior covariance matrix, whose hyperparameters are learnt automatically from the data via evidence optimization (also known as empirical Bayes, or Type II maximum likelihood in the literature\cite{casella_introduction_1985}). These algorithms thus provide STRF estimates with the desired properties even with little or noisy data. Although elegant in theory, the evidence optimization step of this framework is computationally costly: the cubic complexity in the dimensionality of the parameter space renders it difficult to scale to high-dimensional settings as often encountered in sensory neuroscience \cite{wu_complete_2006}. Also, current popular implementations of MAP estimates work mostly under the Linear-Gaussian encoding model, which assumes additive Gaussian noise to the neural response, and are restricted to simple one filter models. 

\begin{figure}
    \centering
        \includegraphics[width=\textwidth]{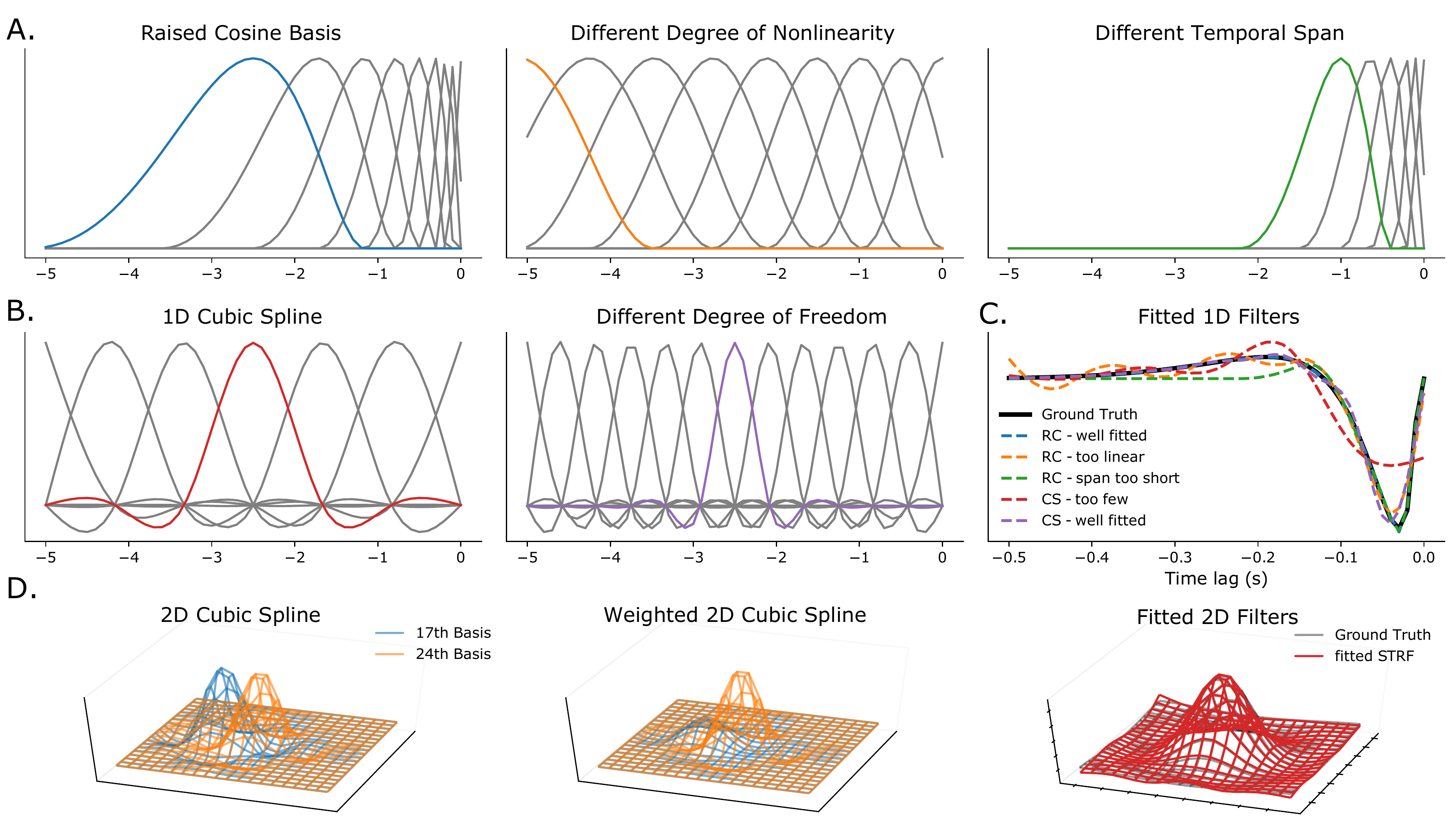}
        \caption{Illustration of the raised cosine and natural cubic spline basis. \textbf{A}: A typical raised-cosine bases as in \cite{pillow_spatio-temporal_2008} controlled by several parameters, such as number of basis functions, the degree of nonlinearity and the placement of endpoints. The choice of these parameters determines the final goodness of fit. \textbf{B}: An equally spaced 1D natural cubic spline, controlled by only one parameter: the number of basis functions (aka the degree of freedom). \textbf{C}: Fitted 1D filters using different bases and different sets of parameters. \textbf{D}: An example of 2D natural cubic splines. Left: Two selected bases of 2D natural cubic splines. Middle: Weighted bases after fitted to a simulated 2D STRF. Right: An example of a fitted 2D STRF overlaid on the ground truth.}
        \label{fig:Figure_basis_example}
    \end{figure}

An alternative approach to encoding prior knowledge for estimating STRFs is to parameterize the STRF with a set of basis functions with the desired properties. In this way, the model fitting becomes much more efficient as the number of parameters to be optimized is significantly reduced, and some degree of smoothness is automatically enforced. Previous studies used a raised-cosine basis (Figure \ref{fig:Figure_basis_example}A) for 1D retinal ganglion cell STRFs in macaque monkey \cite{pillow_spatio-temporal_2008} and Gaussian spectral or Morlet wavelet basis for 2D auditory STRFs in ferrets \cite{thorson_essential_2015}. Those basis functions are chosen to capture certain STRF properties in their corresponding studies, thus they are relatively inflexible to generalize to other settings. Besides, they are typically controlled by multiple hyperparameters, which creates a burdensome and time-consuming model selection process.

Here, we propose to use an alternative basis for receptive field estimation: natural cubic splines (Figure \ref{fig:Figure_basis_example}B). These are known as the smoothest possible interpolant and can be easily extended to high-dimensional settings with minimal assumptions and few hyperparameters \cite{wood_generalized_2017}. This spline basis can be incorporated into a wide range of existing receptive field models (Figure \ref{fig:models_illustration}), such as the Linear-Gaussian (LG) and the Linear-Nonlinear-Poisson (LNP) model\cite{pillow_prediction_2005, pillow_spatio-temporal_2008} for single STRF estimation as well as Linear-Nonlinear-Linear-Nonlinear (LNLN) cascade models\cite{mcfarland_inferring_2013, liu_inference_2017} and spike-triggered clustering methods \cite{liu_inference_2017, shah_inference_2020} for subunit estimation. When combined with proper regularization, smooth and sparse STRFs can be retrieved with very little data and in a computationally efficient manner, while their predictive performance reaches state-of-the-art level. Furthermore, we introduce diagnostic tools for assessing the quality of an estimated STRF: (1) a significance test based on the analytical solution of the STRF covariance matrix and (2) a permutation test based on model prediction. We provide all developed methods as part of the \textsc{RFEst} Python toolbox.

\section{Results}

\begin{figure}[htp]
        \centering
        \includegraphics[width=\textwidth]{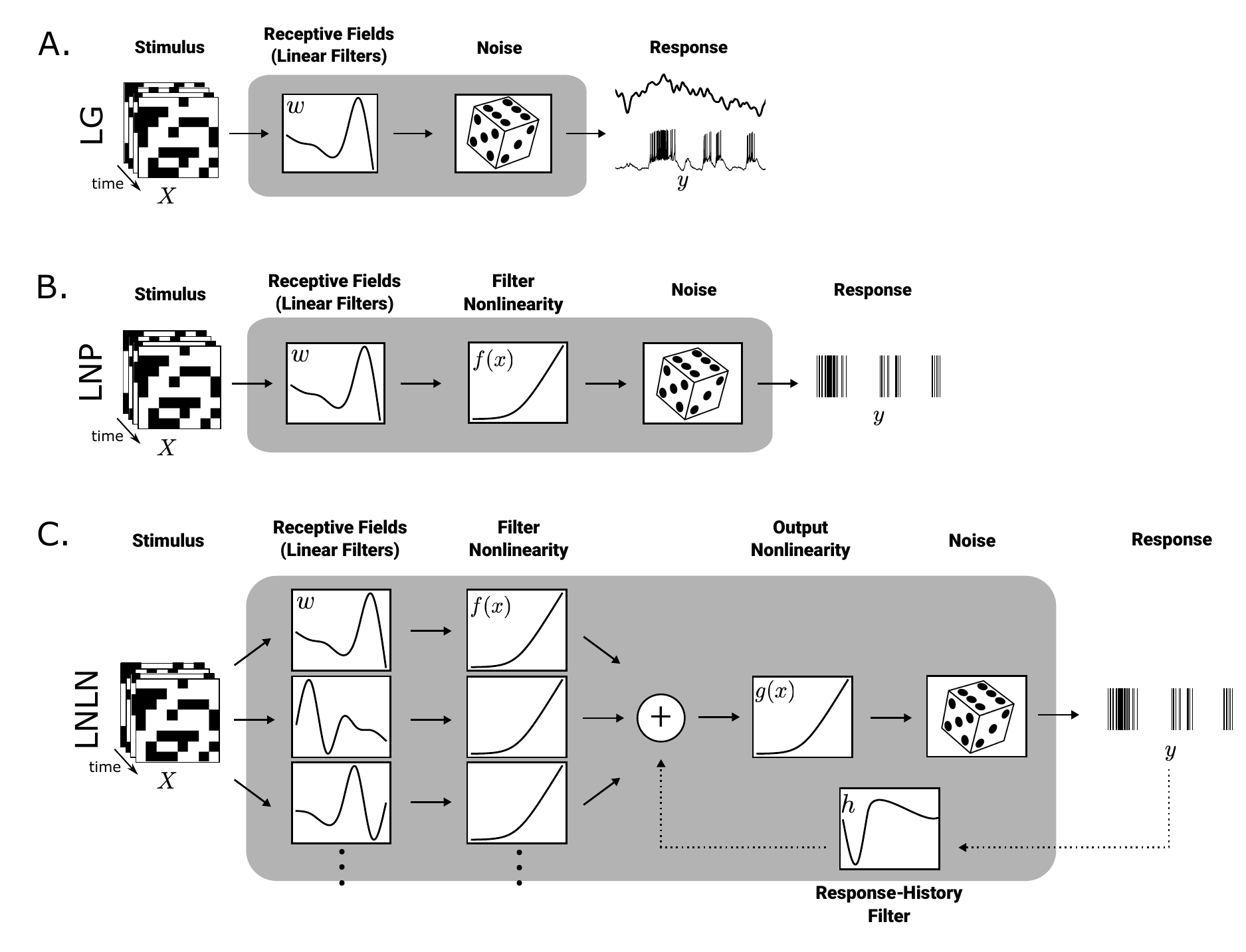}
    
        \caption{Overview of the different spatio-temporal receptive field (STRF) models used in this study. \textbf{A}: Linear-Gaussian (LG) encoding model. $y = Xw + \epsilon, \epsilon \sim N(0, \sigma^2)$, where $X$ is the stimulus design matrix, $y$ the response, $w$ the STRF. \textbf{B}: Linear-Nonlinear-Poisson (LNP) model. $y = f(Xw), y \sim Poisson$, where $f(x)$ is the filter nonlinearity. \textbf{C}: Linear-Nonlinear-Linear-Nonlinear-Poisson (LNLN) model. $y = g(\sum f(Xw)), y \sim Poisson$, where $g$ is the output nonlinearity. Optionally, an autoregressive term convolving previous response with a response-history filter can be added before the nonlinearity to improve the model predictive performance.}
        \label{fig:models_illustration}
    
    \end{figure}

We developed the \textsc{RFEst} Python toolbox for spline-based STRF estimation in various models, including the single-filter LG and LNP models as well as the multi-filter LNLN cascade model (Figure \ref{fig:models_illustration}). We measured the performance of the spline-based methods and compared it with their non-spline versions or other previous methods, using simulated data as well as neural recordings from the retina and the visual cortex. To quantify the goodness of fits for the simulated benchmark data, we measured the mean squared error (MSE) between the ground truth STRF and the estimated STRFs for various methods and studied their dependence on the amount of available data. For the experimental data, we computed the correlation between the predicted and measured responses on a held-out test set as a performance measure. 
 
\subsection{Estimating receptive fields from simulated LG, LNP and LNLN models} 

First, we simulated the responses of a sensory neuron $y$ with additive Gaussian noise using a STRF with 30 time frames and 40 pixels in space in response to white and pink noise stimuli for various stimulus lengths (see Methods \ref{datasets} for details). First, we fit LG models (Figure \ref{fig:models_illustration}A) to these responses using various STRF estimation techniques. 

\begin{figure}
        \centering
        \includegraphics[width=\textwidth]{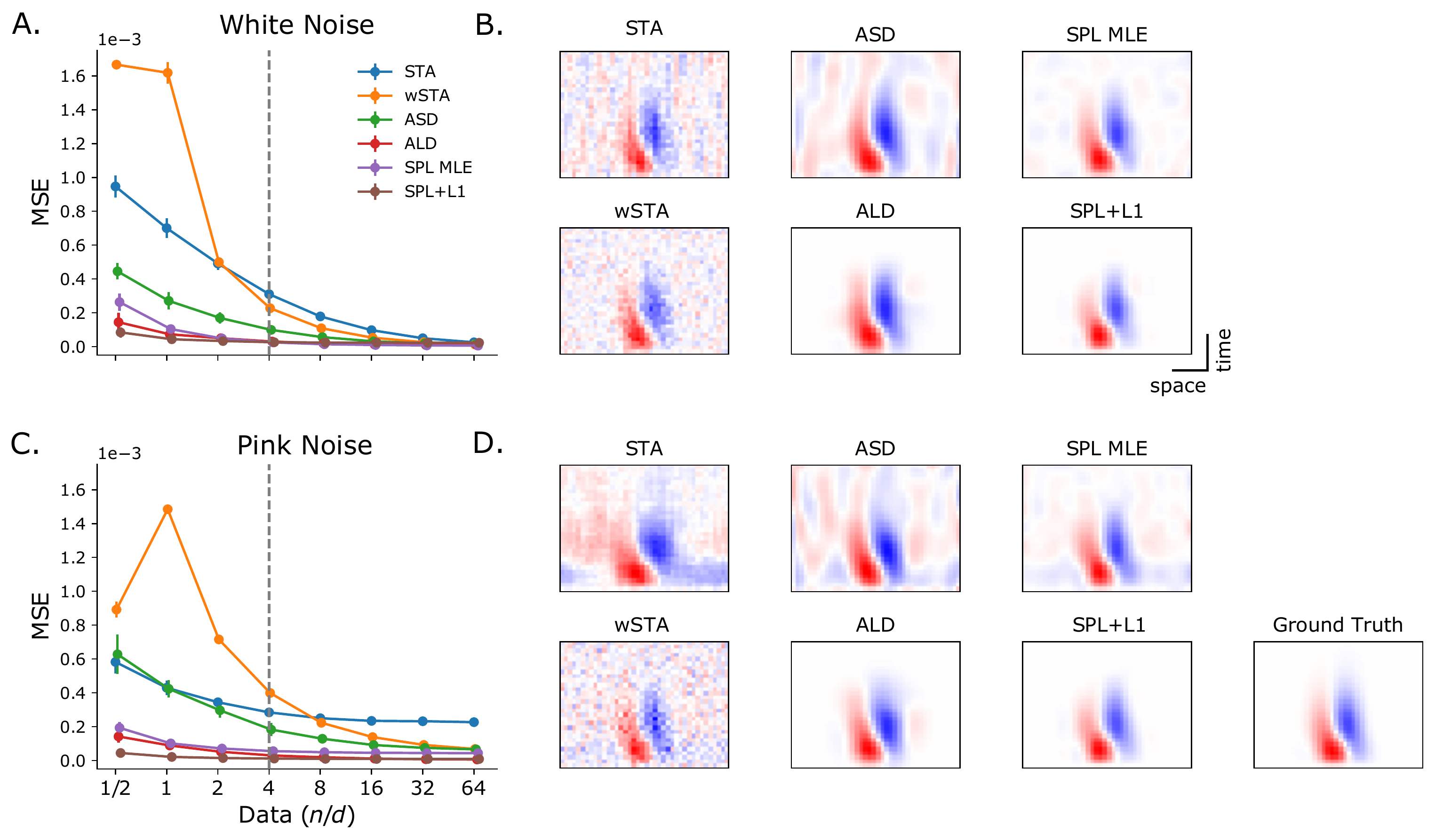}
    
        \caption{Linear-Gaussian (LG) Model. 2D simulated RF stimulated by white and pink noise in various lengths. \textbf{A} and \textbf{C}: Similarity between ground truth and estimated RFs measured as mean squared error (MSE) as a function of the amount of simulated data. Error bars are computed by averaging over 10 trials with different random seeds. \textbf{B} and \textbf{D}: Ground truth and examples of estimated RFs by different methods at n/d=4, indicated by vertical grey dashed line in \textbf{A} and \textbf{C}.}
        \label{fig:Figure_Sim_LG}
    
    \end{figure}

We measured the MSEs between the ground truth STRF and the different STRF estimates with respect to the ratio of the number of samples $n$ to the number of parameters $d=30\times40$, averaged over 10 different random seeds (Figure \ref{fig:Figure_Sim_LG}A). 
As expected, the STA and wSTA approached the ground truth only in high data-to-parameters regimes. For limited data (e.g. n/d=4), the estimated STRF showed substantial background noise (Figure \ref{fig:Figure_Sim_LG}B). In the case of pink noise, the STA was in fact never able to recover the STRF due to the high correlation between pixels in the stimulus \cite{paninski_convergence_2003} (Figure \ref{fig:Figure_Sim_LG}C-D). 

We also applied more advanced techniques to estimate the STRF to the simulated data, such as automatic smoothness determination (ASD, aka MAP with smoothness prior \cite{sahani_evidence_2002}) and automatic locality determination (ALD, aka MAP with locality prior \cite{park_receptive_2011}). These estimates were much closer to the ground truth already with a relatively small amount of data, with ALD outperforming ASD for both white and pink noise (Figure \ref{fig:Figure_Sim_LG}). However, their computation time for estimating the optimized prior was long due to the high computational complexity. 

We then applied a spline-based version of the wSTA (SPL wSTA) with and without L1 regularization on the basis function coefficients (see Methods). This simple change of basis functions led to a STRF estimator which performed very well for all n-d-ratios tested, even better than ASD and ALD, especially when using L1 regularization (Figure \ref{fig:Figure_Sim_LG}). At the same time, this estimator was also computationally very efficient, as the number of coefficients to be estimated was reduced to the number of basis functions $b$ (for this example, the STRF had $d=30\times40=1200$ parameters, whereas the number of basis functions was only $b=9\times12=108$) 

We next simulated responses $y$ from an LNP model (Figure \ref{fig:models_illustration}B) with Poisson noise similar to before using white and pink noise for stimulation. For simplicity, we used an exponential nonlinearity and treated it as known during the estimation procedure. We estimated the linear filter from different amounts of data using the STA, wSTA, ASD and ALD, as well as spline and non-spline LNP fitted via gradient descent with L1 regularization (referred to as LNP+SPL and LNP, respectively). Note that the original ASD and ALD were not designed for the LNP but the LG model (though some progress has been made for the former \cite{park_bayesian_2013}). Despite this model mismatch, they still recovered good estimates of the STRF (as has been demonstrated in the original papers \cite{sahani_evidence_2002, park_receptive_2011}).

\begin{figure}[htp]
        \centering
        \includegraphics[width=\textwidth]{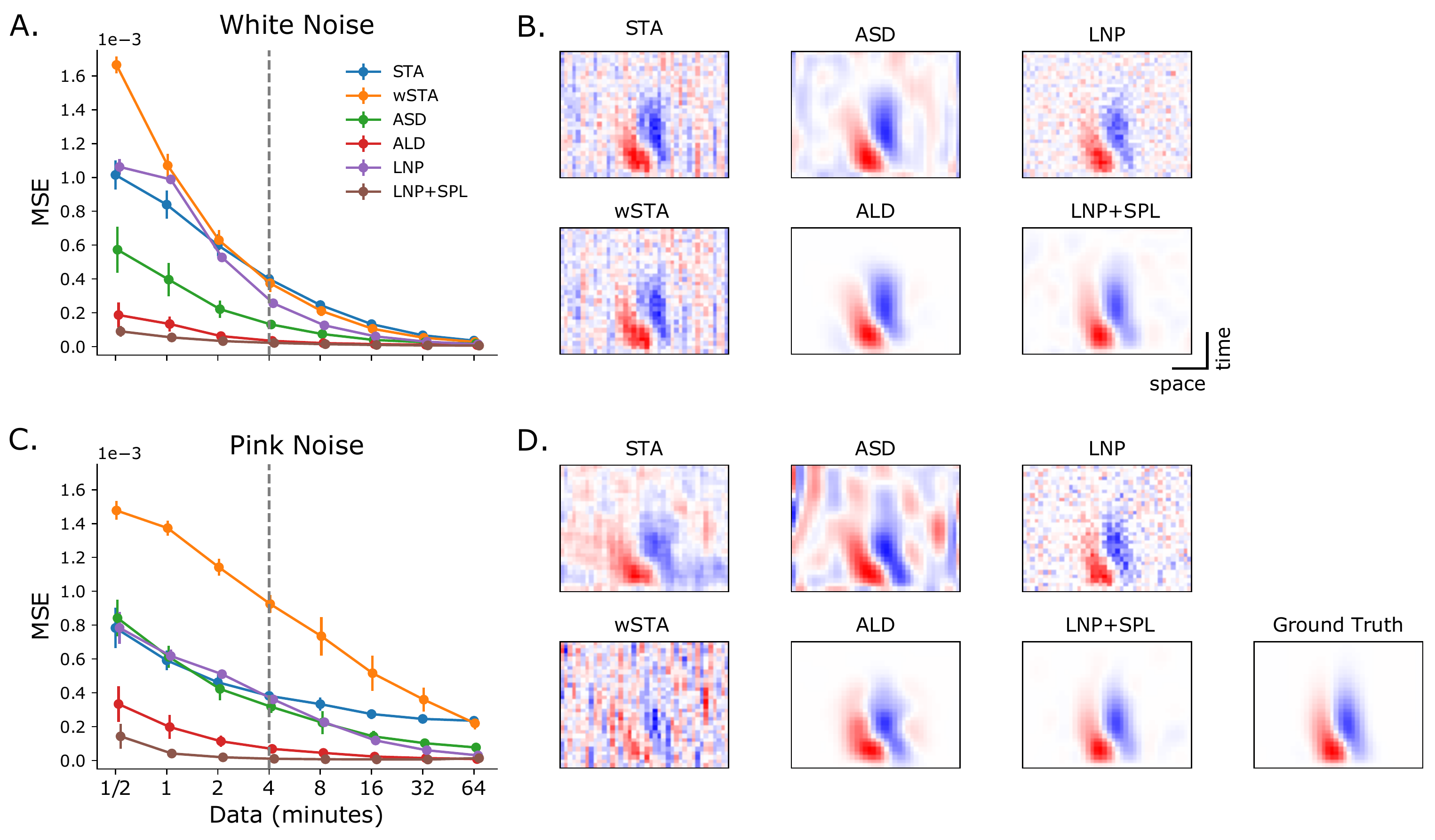}
    
        \caption{Linear-Nonlinear-Poisson (LNP) Model. A simulated 2D RF stimulated by white and pink noise for various stimulus lengths. \textbf{A} and \textbf{C}: Details are the same as in Figure \ref{fig:Figure_Sim_LG}. \textbf{B} and \textbf{D}: Example estimated RFs by different methods at data=4 minutes, indicated by vertical grey dash line in \textbf{A} and \textbf{C}.}
        \label{fig:Figure_Sim_LNP}
    \end{figure}

Again, we found that different methods resulted in estimates of different quality depending on the available amount of data (Figure \ref{fig:Figure_Sim_LNP}). Here, in contrast to the LG model, we changed the x-axis of Figure \ref{fig:Figure_Sim_LNP}A and C to time in minutes for better presentation, as the model performance was dependent on the number of effective samples (frames that contain at least one spike) instead of the total number of samples, and the stochastic nature of the Poisson spike generator yielded different number of spikes for each trial. For both stimulus conditions, we found that the spline-based LNP yielded extremely data-efficient estimates of the linear filter compared to all other methods. In contrast, gradient-based optimization of an LNP model without spline basis, even with L1 regularization, performed only slightly better than the STA and wSTA computed under the mismatched LG model assumptions. The gap in performance of the LNP+SPL and MAP estimators of ASD and ALD were larger here compared to the LG case. This was partially due to the fact that the number of effective samples was smaller in Poisson spiking data, but the mentioned model mismatch between the cost function of the LG model optimized by the ASD and ALD methods and the data likely also contributed.

We finally investigated how recently proposed subunit models (Figure \ref{fig:models_illustration}C), in particular an LNLN cascade model \cite{mcfarland_inferring_2013, maheswaranathan_inferring_2018} and spike-triggered clustering or non-negative matrix factorization methods\cite{shah_inference_2020, liu_inference_2017}, can be made more data-efficient using a spline basis. We compared their performance with and without spline augmentation (see Methods \ref{method-spike-triggered-clustering}). The assumption behind the spike-triggered clustering methods is that each spike can be attributed to a specific subunit of a STRF, so that each subunit can be viewed as a STA of a cluster of spikes. If this assumption holds, then a spline-approximated STA should be able to serve as a better centroid for clustering, as the smoothed STA is closer to the ground truth subunit STRF. We used k-means clustering as an in-place of the previous published soft-clustering method \cite{shah_inference_2020} and used a multiplicative update rule\cite{ding_convex_2010} for semi-nonnegative matrix factorization (semi-NMF) \cite{liu_inference_2017}. Moreover, we modified the previous published semi-NMF \cite{liu_inference_2017} and imposed the nonnegative constraint to the weight matrix instead of the subunit matrix, which allowed us to retrieve antagonistic STRFs. 

    \begin{figure}[t!]
        \centering
        \includegraphics[width=\textwidth]{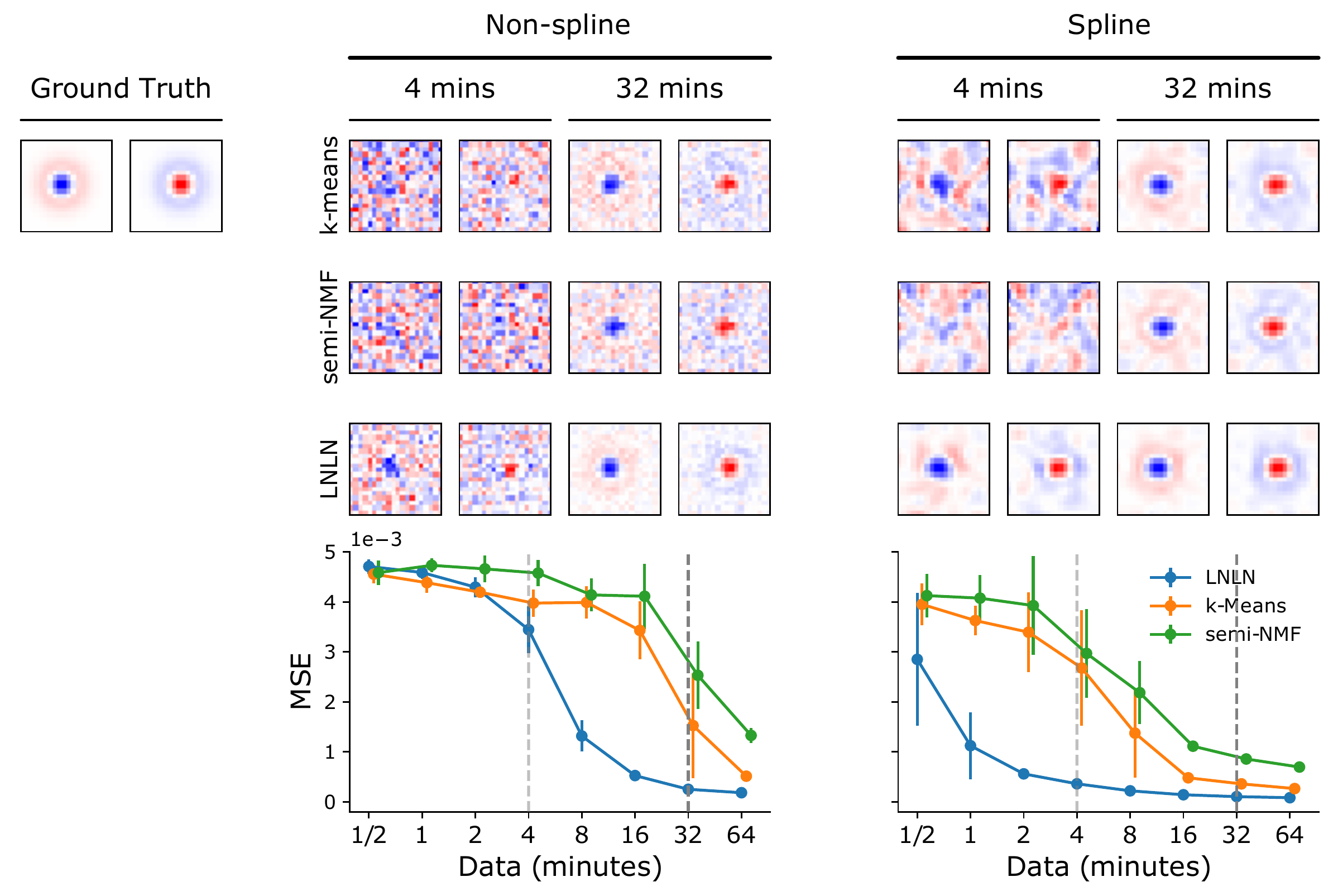}
        \caption{Subunit models. Two antagonistic difference-of-Gaussian 2D RFs stimulated by white noise. The subunit STRFs were estimated by k-means clustering, semi-NMF and LNLN, without and with splines. Examples shown are estimated subunits by different methods at data = 4 and data = 32 minutes, indicated by vertical light and dark grey dash line in the bottom panel, respectively. Similarity between ground truth and estimated RFs measured as mean squared error (MSE) as a function of the amount of simulated data.}
        \label{fig:Figure_Simulation_lnln}
    \end{figure}

We simulated responses $y$ with Poisson noise from the model by stimulating two 20x20 pixels antagonistic center-surround subunits with different lengths of white noise stimuli. Both filter outputs first went through an exponential nonlinearity, were then summed up, and finally went through a softplus output nonlinearity.

We estimated subunit STRFs using k-means clustering or semi-NMF of the spike-triggered ensemble, and maximum-likelihood estimation of the LNLN model directly, with or without a spline basis (Figure \ref{fig:Figure_Simulation_lnln}). Similar to the results from previous simulations, retrieved subunit STRFs from spline-based models were generally much more similar to the ground truth than models without spline, and required a smaller amount of data to achieve a similar level of performance. For example, when using a spline basis, we were able to retrieve the two subunit filters at high quality using only 4 minutes of data. With this amount of data, all other methods still showed essentially only noise with a hint of signal in the estimate of the linear filter. 

It is worth noting that the computational efficiency of LNLN estimation was improved a lot by using a spline basis, as the number of coefficients to be estimated was significantly reduced. In contrast, the computation time for both spike-triggered clustering methods was actually increased by incorporating splines, because more computational steps were added to the original algorithms in those modified versions (see Methods \ref{method-spike-triggered-clustering}).

\subsection{Comparing computation time between different methods and model selection}

It was not straightforward to fairly compare the computation time between MAP-derived and spline-based GLMs, as the most time-consuming part for MAP was to estimate the prior hyperparameters via evidence optimization, while spline-based GLMs need to resort to cross-validation for selecting the number of basis functions ($b$, also known as the degrees of freedom) for each dimension. Yet, we can still get an impression of the algorithmic efficiency of each method by profiling their most time-consuming step. 

\begin{figure}
        \centering
        \includegraphics[width=\textwidth]{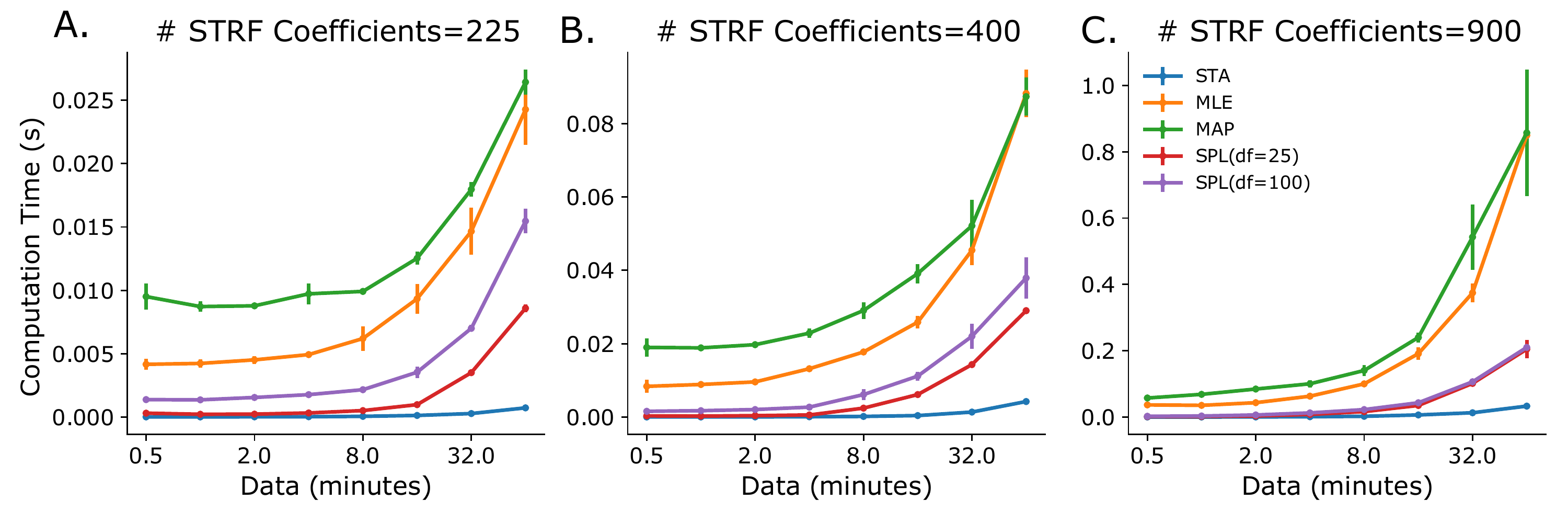}
    
        \caption{Comparison of computation time between different methods with different number of STRF coefficients, amount of data and number of spline basis. MAP refers to both ASD and ALD methods, as the size of the covariance matrix was the same for both techniques. Measurements of computation time were averaged over 10 repetitions and performed on a 2019 16-inch MacBook Pro with a 2.3 GHz 8-core Intel Core i9 and 16 GB of RAM. See Method \ref{method-computation-time} for more information.}
        \label{fig:time_comparison}
    \end{figure}

Therefore, we compared the time taken by the computation of the STA (see equation \ref{eq_sta} in Methods), wSTA (equation \ref{eq_wsta}), MAP (equation \ref{eq_map}) and SPL wSTA (equation \ref{eq_spl_mle}) for different amounts of data by separately profiling the single most time consuming computation (Figure \ref{fig:time_comparison}). Not surprisingly, STA was the fastest, as it had a linear time complexity ($O(n)$) and its computation time depended mainly on the amount of data ($n$), while wSTA and MAP were the slowest, as they were both limited by the calculation of the inverse matrix, and their computation time grew in a cubic manner with respect to the number of coefficients ($w$) in the STRF due to the cubic time complexity ($O(w^3)$) of the matrix inversion. SPL on the other hand, was less time-consuming than wSTA and MAP, even though the time complexity was still cubic. However, the improvement was due to the reduced number of model parameters by parameterizing the STRF coefficients with the spline basis ($b$ coefficients), which resulted in a much smaller covariance matrix to invert ($O(b^3)$), thereby speeding up the most time-consuming step of the algorithm substantially. Thus, spline-based STRF estimators offer the best available compromise between STRF estimation quality and computation time.

The total amount of time spent on finding an optimal STRF depends on many factors, such as how the model is initialized, the choice of learning rate and stopping criteria for gradient descent optimization. Importantly, the way hyperparameters are selected also contributes significantly: most commonly, the model prediction performance is monitored for on a held-out data set over a discrete grid of hyperparameters. Even for evidence optimization methods, in which the main selling point is to learn the optimal hyperparameters automatically, cross-validation can still be used to assess model goodness of fit and generalization performance \cite{sahani_evidence_2002, park_bayesian_2011}. For spline-based GLM, this requires refitting the same model as often as the number of hyperparameter sets. Given that splines are equally spaced, a good strategy to accelerate the process would be to measure the performance of SPL wSTA over a grid of possible candidates from very few to many, choose the number when the performance is starting to plateau, then fit SPL+L1 with gradient descent to retrieve a sparse STRF.

\subsection{Diagnostic tools}
\label{section-diagnostic-tools}
    \begin{figure}[htp]
    
        \centering
        \includegraphics[width=0.825\textwidth]{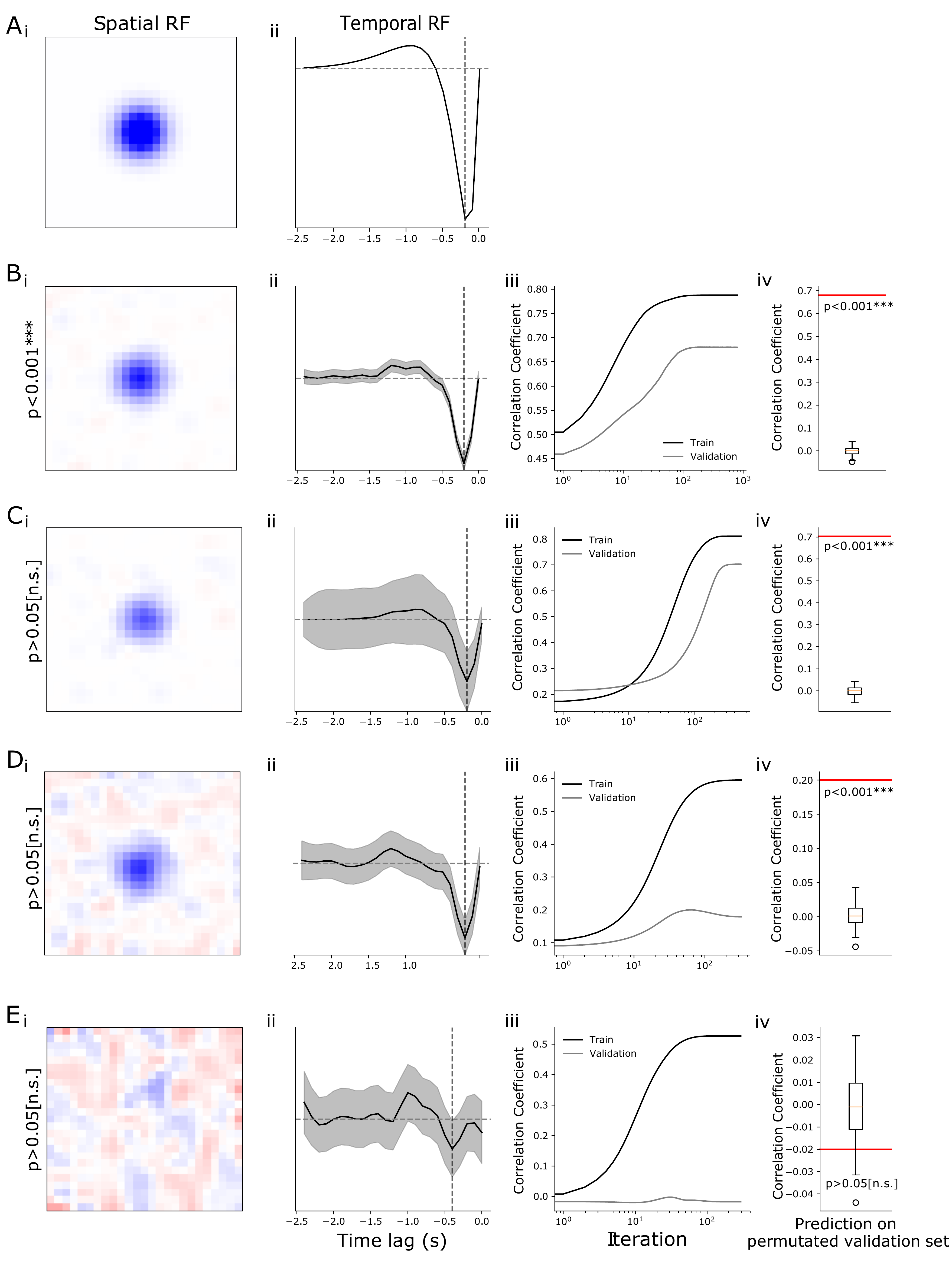}

        \caption{Examples of diagnosing STRF quality with confidence interval, p-value, prediction performance. \textbf{A}: Ground truth STRF with the dimensions 25x25x25. The 3D STRF is the Kronecker product of the respective temporal and spatial RF. \textbf{B}: A well fitted STRF with the sample size equals to $d$. (\textbf{i}) the estimated spatial RF at the minimum of temporal RF, indicated by the gray vertical dotted line in (\textbf{ii}); (\textbf{ii}) the estimated temporal RF; (\textbf{iii}) the change of correlation between the data and predicted response over all gradient descent iteration. (\textbf{iv}) the predicted performance on the permuted validation set. \textbf{C}: Same as \textbf{B} but with half amount of data. \textbf{D}: Same as \textbf{B} but with stronger additive Gaussian noise. \textbf{E}: Same as \textbf{B} but the simulated response were permuted.}
    \label{fig:Figure_diagnostics}
    \end{figure}

In addition, even if a well-regularized model is chosen according to predictive performance, additional diagnostic tools can be useful to decide whether a certain STRF is reasonable and shows a clear structure (Figure \ref{fig:Figure_diagnostics}). We devised three diagnostic tools that can be employed to this end: (1) the confidence interval (CI) of the fitted STRF and its corresponding p-values based on Wald statistics (with the null hypothesis that all STRF coefficients are zero, see also Method \ref{method-ci}); (2) the difference between the metrics of the training and the validation set over all iterations of the optimization; and (3) a permutation test on model prediction for validation and/or test data set (Method \ref{method-permutation}). 

To demonstrate their usefulness, we investigated a ground truth 3D STRF under a Linear-Gaussian model. We showcased four different scenarios (Figure \ref{fig:Figure_diagnostics}): First, we looked at a well-fitted STRF (Figure \ref{fig:Figure_diagnostics}A). Here, the CI was narrow and significantly different from an all-zero STRF (p<0.001), the gap between the training and validation prediction metrics was small and the prediction on the validation set was better than chance (p<0.001). The confidence interval and its corresponding p-value were determined by the training data. Next, we looked at scenarios where the amount of data was small or the signal-to-noise ratio (SNR) was low (Figure \ref{fig:Figure_diagnostics} C and D). Here, the STRF did not pass the significance test due to a lack of power, but it is still possible that the model predictions on the validation set were better than chance. Additionally, if the gap between the performance metric on the training and validation sets gets too large, it would be a good idea to refit the model with a stronger regularization. Finally, in the case of the data with extremely low SNR (Figure \ref{fig:Figure_diagnostics}E), which we generated by randomly permuting the simulated response, the prediction performance of the fitted STRF was just at the chance level regarding all indicators. 

\subsection{Application to experimental data}

We next documented the complete process of using gradient-based estimation for the parameters of spline-based LNP and LNLN models in a previously published experimental data set: extracellularly recorded spikes from tiger salamander retinal ganglion cells (RGC), stimulated by 3D non-repeat checkerboard white noise \cite{liu_inference_2017} (Figure \ref{fig:Figure_Expdata}A). In addition, we applied the same techniques to 2-photon calcium imaging data recorded from a mouse retinal ganglion cell soma stimulated by checkerboard white noise (Figure \ref{fig:Figure_Expdata}B) (using the positive gradient of the calcium trace as the response). We compared the results of spline-based models with non-spline versions.

For both data sets, only the first 10 minutes of the whole recording were used for fitting the models, another 2 minutes were used for model validation and another 4 sets of 2 minutes for testing. The predictive performance was measured by the average of the four correlation coefficients of the predicted and measured responses. We set the number of subunits $k=4$ throughout. The dimensionality of the 3D STRFs were [20, 25, 25] and [25, 15, 15], respectively. We added diagnostic information as in Figure \ref{fig:Figure_diagnostics}. Moreover, a contour was drawn on the spatial filter to illustrate the spatial extent of the STRF. 

We fitted the non-spline LNP and LNLN with different L1 regularization weights from 0 and 10, with steps of 1. 

For spline LNP and LNLN, the number of basis functions (aka degree of freedom, df) for temporal and spatial dimensions (assuming x and y dimensions here used the same df) were first selected based on the predicted performance of the maximum likelihood estimates on the validation set (from 1/3 to 2/3 of the pixels in each dimension, with steps of 1). Finally, for data set (1), the grid search took 11.2 minutes on the same hardware setup used for Figure \ref{fig:time_comparison} and yielded the optimal df=[9, 9, 9] ; for data set (2), it took 2.4 minutes and yielded the optimal df=[12, 9, 9] (Figure \ref{fig:Figure_expdata_cv}). See also how different amounts of training data affects the optimal df in Supplementary Section 1.

    \begin{figure}[htp]
    
        \centering
        \includegraphics[width=0.8\textwidth]{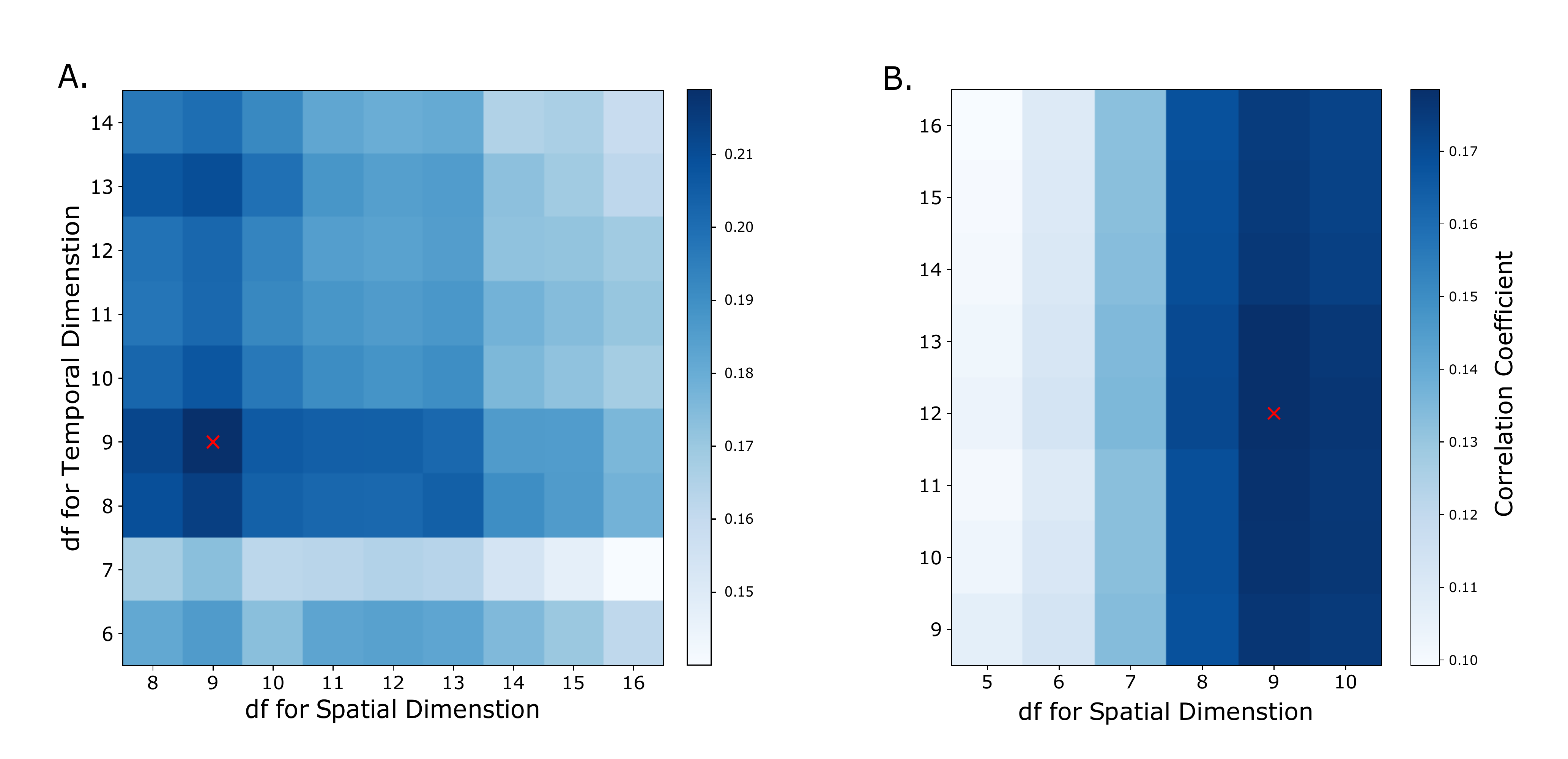}

        \caption{Choosing the number of basis functions with cross-validation. The predictive performance of the SPL wSTA on the validation set for a grid of different degrees of freedom (df). The optimal df are indicated by red crosses. \textbf{A}: Spike recordings from a salamander RGC. \textbf{B}: 2-photon calcium imaging recordings from a mouse RGC (see Methods).}
    \label{fig:Figure_expdata_cv}
    \end{figure}

Then, we fitted spline LNP and LNLN with L1 regularization weights ranged from 0.5 and 1.5, with steps of 0.1 (as we observed, the spline model required a smaller value for regularization due to the reduction of model parameters). The ordered grid search for the L1 weight, starting with the smallest value, was interrupted if the current predictive performance was lower than the previous one. All models were initialized with maximum likelihood estimates (plus small additive Gaussian noise) and fitted by gradient descent with 1500 iterations, but an early stop was also deployed with two triggers: either the training cost changed less than 1e-5 for the last 10 iterations, or the validation cost monotonically increased for the last 10 iterations. Sometimes the optimization didn't stop after reaching the lowest validation cost due to the adaptive learning rate resulting in the cost changing in a zigzag manner. But regardless of the stopping trigger, the model at the lowest validation cost was returned. The computation time of each fit for all four models was summarized by box plots in Figure \ref{fig:Figure_Expdata}A$_{iv}$ and B$_{iv}$. 

Fitting a 3D STRF with gradient descent directly was considered infeasible unless certain regularization was imposed \cite{maheswaranathan_inferring_2018}, and the situation would become even worse for subunit models, as the number of model parameters increases linearly with the number of subunits. Previously methods \cite{liu_inference_2017, shah_inference_2020} (as shown in Figure \ref{fig:Figure_Simulation_lnln}) relied on dimensionality reduction techniques to retrieve subunit STRFs but those methods could not be applied to 2-photon calcium imaging data directly without an extra step of spike inference. Spline GLMs not only overcome both of these problems but also allow us to assess the quality of the estimated STRF with CI and their corresponding p-values (In principle, the same statistics can also be applied to non-spline GLMs, yet a larger number of model coefficients would result in high computational cost for computing the STRF posterior covariance).   

We found that for both example neurons from the two data sets, STRFs estimated by non-spline models were noisy and prediction performance was only moderate (Figure \ref{fig:Figure_Expdata}A$_{i}$, B$_{i}$, A$_{iii}$, B$_{iii}$). This was especially prominent for LNLN estimates, as here the number of parameters to estimate exceeded the amount of available data to fit the models (Figure \ref{fig:Figure_Expdata}A$_{ii}$ and B$_{ii}$). In contrast, the spline-based LNP and LNLN estimated in general were smooth and sparse even with a severely limited amount of data, and they consistently outperformed their non-spline versions (Figure \ref{fig:Figure_Expdata}A and B).

The number of subunits that could be retrieved by the model was limited by the amount of available data (total length of the recording, and for spike data, the total number of spikes) as well as the signal-to-noise ratio of the data. With the limited data in our examples, we found three and two non-overlapping subunits for the two neurons, respectively (Figure \ref{fig:Figure_Expdata}A$_{ii}$ and B$_{ii}$). Additional subunits available to the model were pushed to near zero by the L1 regularization and the significance test failed to reject the null hypothesis that all coefficients in those subunit STRFs were close to zero. But as noted in the previous section, the p-value should not be the only judge for the quality of the estimated STRF. As shown in \ref{fig:Figure_Expdata}B, the STRF from the spline LNP model was also non-significant but looked reasonable from the snapshot visualization. In cases like this, other tools such as the permutation test shown in the previous section would be required to determine whether the STRF in question was good or not.

We provided two more examples of visual neurons (RGC and V1) and one example of non-visual neuron (entorhinal grid cell) in Supplementary Section 2.

    \begin{figure}
        \centering
        \includegraphics[width=0.95\textwidth]{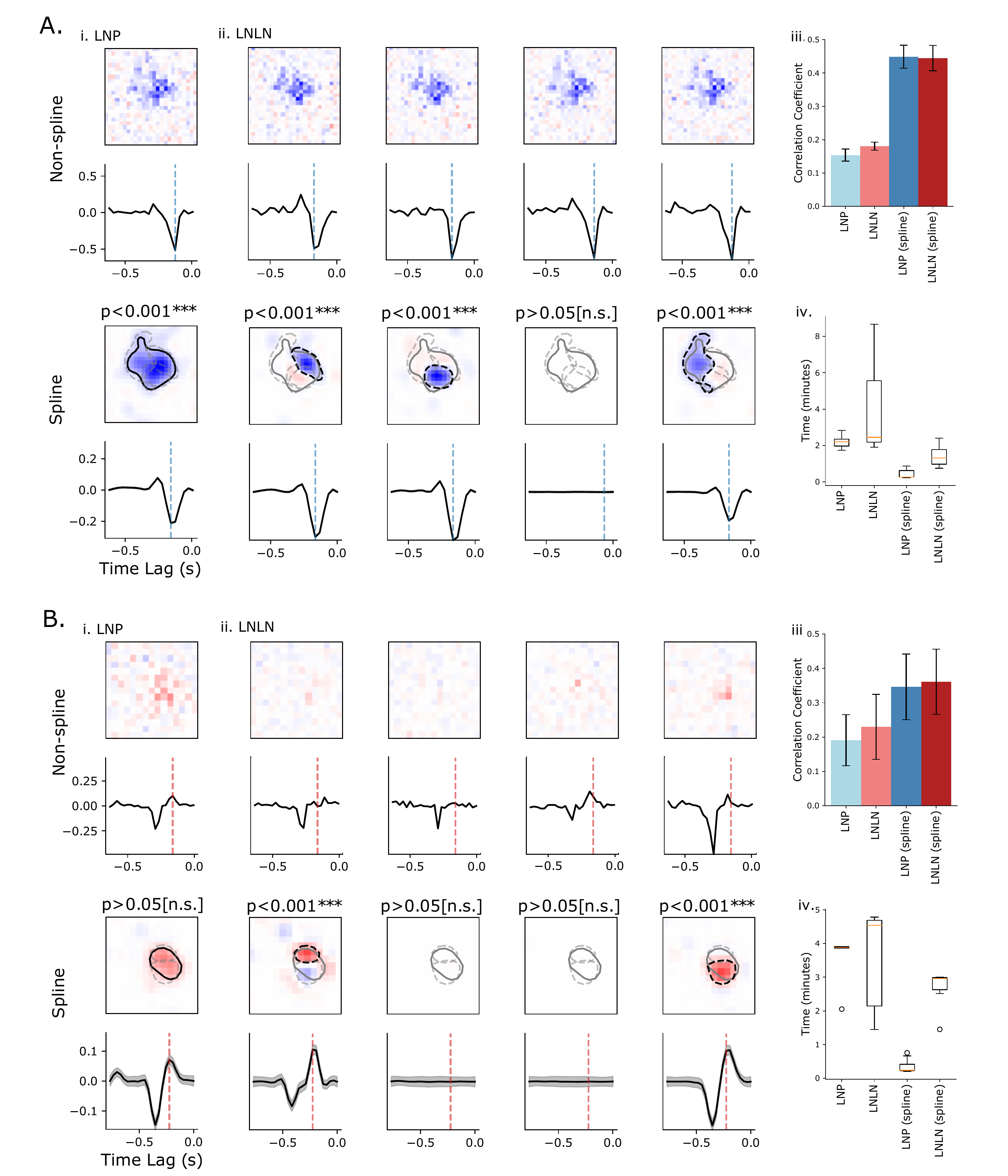}
    
        \caption{Spline-based LNP and LNLN consistently outperformed their non-spline versions on both example neurons from two data sets: \textbf{A}: Spike recordings from a salamander RGC, \textbf{B}: 2-photon calcium imaging data from a mouse RGC. (see Methods \ref{datasets} Experimental data for details). \textbf{i} and \textbf{ii}: STRFs estimated from LNP and LNLN with or without spline, where number of subunits $k=4$ for all LNLN models; Shading for temporal filters show 95\% confidence interval (see Methods, \ref{method-ci}), which is too small in \textbf{A} to be visible. \textbf{iii}: predictive performance comparison of LNP and LNLN with and without spline, measured by the averaged correlation coefficients of the test set response and the predictive ones. Error bars were computed from 4 sets of test data. \textbf{iv}: Computation time of each fit for the same models were summarized in the box plots.}
        \label{fig:Figure_Expdata}
    \end{figure}

\section{Discussion}

Here, we showed that natural cubic splines form a good basis for STRF estimation by tackling one of the main challenges in sensory neuroscience: to efficiently estimate high-dimensional STRFs with limited data. Using this basis function set, we showed that inference for single filter LN models with different noise models as well as LNLN cascade models can be efficiently performed from both spiking as well as two-photon imaging data, even if the signal-to-noise ratio is low (and see also how spline-based methods compare to previous methods in Table \ref{tab:model_comparison}). We provide an implementation of the proposed methods in the \textsc{RFEst}-package for Python. 

\begin{table}
    \centering
    \begin{tabular}{llllll}
    \hline
    Methods  & Stimuli Type     & Noise Distribution & Data needed & Complexity        & Multiple Filters \\ \hline
    STA      & White Noise only & Gaussian           & a lot       & $O(n)$            & No               \\
    wSTA     & Any              & Gaussian           & a lot       & $O(w^3)$          & No               \\
    MAP      & Any              & Gaussian           & little      & $O(w^3)$          & No               \\
    LNLN     & Any              & Poisson            & a lot       & $O((wk)^3)$       & Yes              \\ 
    SPL LG   & Any              & Gaussian           & little      & $O(b^3)$          & No               \\
    SPL LNP  & Any              & Poisson            & little      & $O(b^3)$          & No               \\
    SPL LNLN & Any              & Poisson            & little      & $O((bk)^3)$       & Yes              \\ \hline
    \end{tabular}
    \caption{Model comparison. Here, $n$ is the number of samples, $w$ the number of STRF coefficients, $b$ the number of spline basis coefficients and $k$ the number of subunits. In most cases, $b \ll w$.}
    \label{tab:model_comparison}
\end{table}

\subsection{Choice of basis functions}

We chose the natural cubic spline basis to represent the linear filters as default in our toolbox, because among many other spline bases such as cyclic cubic splines, B-splines, or thin-plate splines, natural cubic splines have been shown to be the smoothest possible interpolant through any set of data and theoretically near-optimal to approximate any true underlying functions closely \cite{wood_generalized_2017}. Conveniently, assuming all knots are spaced equally in each dimension, natural cubic splines only require setting one hyperparameter: the degree of freedom, or the number of basis functions. This is unlike the raised-cosine basis used previously\cite{pillow_spatio-temporal_2008}, which is governed by multiple hyperparameters. Also, natural cubic splines can be easily extended to high-dimensional settings, also known as tensor product bases, by simply taking the row-wise Kronecker product of the basis matrix along each dimension (see Method \ref{method-cubic-spline}), making them uniquely suitable for estimating STRFs. There is also an interesting connection between smoothing splines and Gaussian processes. As shown by \cite{kimeldorf_correspondence_1970, bay_generalization_2016, bay_generalization_2016}, spline regression converges to the MAP solution of a Gaussian process regression under certain constraints. 

Through extensive validation with simulated and real experimental data sets, we showed that STRFs estimated with this basis can not only be more accurate than previous state-of-the-art methods, but the estimation procedure was also computationally much more efficient. The estimated STRFs were smoothed automatically, as they were the linear combination of a set of smooth basis functions. Further, sparsity and localization could often be achieved by adding an L1 regularization term to the loss function, which pushed the coefficient for the less relevant basis functions to zeros. Moreover, by parameterizing STRFs with basis functions, the number of coefficients could be significantly reduced, hence model fitting became much more computationally efficient, and the amount of data needed for the optimization to converge was also significantly less.  

\subsection{Potential pitfalls of using splines in non-ideal situations}

One caveat of using splines to fit STRFs is that some STRFs might not need smoothing, thus using splines would lead to suboptimal results. For example, if a STRF occupies only one pixel in space or the amplitude changes sharply in the temporal profile, spline-based model would likely yield suboptimal results by yielding overly smoothed filters. However, one can identify these cases by monitoring the cross-validation performance on held-out data while increasing the number of basis functions in each dimension. If the performance keeps increasing as the number of basis functions reaches the number of pixels in each dimension, a spline-free model could be preferable.

Another concern of using splines is that it would be difficult to tell a poor fit from a good one in some cases, as a poorly-fitted spline-based STRF would also look like a smooth STRF, unlike pixel-based methods that yield a noisy STRF when the fit is poor. In these cases, the diagnostic tools we introduced in Section \ref{section-diagnostic-tools} and Figure \ref{fig:Figure_diagnostics} would be helpful to assess the quality of the STRFs.

\subsection{Other approaches to efficient estimation of receptive fields}

Similar to our approach, fast automatic smoothness determination (fASD) \cite{aoi_scalable_2017} has been suggested to speed up the process of evidence optimization in ASD. Because the smoothness prior covariance can be represented in frequency coordinates and many high-frequency coefficients of the Fourier-transformed filter tend to be small, the size of the prior covariance matrix can be reduced by cutting off those small value frequencies (an idea 
similar to the modified wSTA \cite{theunissen_estimating_2001}), leading to a speed-up in the optimization process. However, as the efficiency of fASD is controlled completely by the initial frequency truncation, and the level of frequency to be truncated has to be fixed before the evidence optimization starts, one needs to resort back to grid search and cross-validation for choosing a good starting point, making the automatic hyperparameters selection via evidence optimization less automatic and comparable to the need to find the optimal degrees of freedom for the spline basis. Also, as fASD (and original ASD) rely on only a few hyperparameters, it might not even require to do evidence optimization at all but just to find the optimal hyperparameters through cross-validation completely. Importantly, fASD is so far only applicable within the framework of linear models with Gaussian additive noise, while our approach is ready to use in nonlinear models with non-Gaussian noise, as shown for LNP and LNLN models.

Regarding the inference for multiple STRFs in LNLN-cascade models, previous work has attempted to provide efficient algorithms by first retrieving subunit STRFs through spike-triggered clustering methods such as soft-clustering \cite{shah_inference_2020} or semi-NMF \cite{liu_inference_2017}. These were subsequently used to estimate other model parameters via gradient descent while holding the subunit STRFs fixed. Utilizing only the collection of stimuli that elicit a spike, these approaches have advantages in terms of the computational cost in time and memory compared to the direct fitting of the complete model using gradient descent \cite{mcfarland_inferring_2013, maheswaranathan_inferring_2018}. 

The ease of using these algorithms relying on the spike-triggered ensemble under white noise stimulation is also the disadvantage of these approaches -- they can only be applied to spike data under white noise stimulation. In many cases, experiments today record more diverse measurements from neurons, such as two-photon calcium imaging \cite{baden_functional_2016, ran_type-specific_2020} or synaptic current recording \cite{trong_origin_2008} under more diverse types of stimulation, such as correlated noise and natural stimuli \cite{qiu_mouse_2020}. In those cases, spike-triggered analysis can not be used directly without modification, as the spike-triggered average is no longer the maximum likelihood estimate \cite{paninski_convergence_2003}. In contrast, our method does not rely on Gaussianity of the stimulus or the discreteness of the response, and in principle can be generalized to other stimuli distribution and other response types. 

In addition, current spike-triggered clustering methods rely on temporally collapsing the 3D stimuli to further reduce the model complexity, assuming all subunits share the same temporal response kernel. This limits the usability of these methods to some very specific cell types and will not account for the spatio-temporal inseparable computation implemented by neurons receiving inputs from multiple sources, such as bi-stratified On-Off direction-selective cell in the retina \cite{taylor_diverse_2002}. In this case, a more general approach such as our method should be preferred.

\subsection{Further extensions}

Here, we only showcased how to estimate STRFs using a natural cubic spline in a set of popular models with standard gradient descent with L1 regularization. This simple spline add-on can be directly incorporated into other previous models \cite{mcfarland_inferring_2013, maheswaranathan_inferring_2018} with different constraints, regularization and optimization techniques, as it only changed the linear step of the models without modifying other parts. 

Furthermore, to further improve model prediction performance, a response-history filter (as shown in Figure \ref{fig:models_illustration}) and flexible nonlinearity can be added to the model \cite{williamson_equivalence_2015, maheswaranathan_inferring_2018}, as both of which can also be parameterized by a 1D spline basis. With such extra flexibility, a model can better adapt to the cell-specific firing threshold and adaptation to the stimulus ensemble. Still, even with the response-history filter, a STRF model may still not be able to account for the total variability of the neuronal responses, if the response is influenced by other sources. Future extension of the current GLM for STRFs can consider incorporating more diverse sources of behavioral data \cite{balzani_efficient_2020} of the experimental animal under experimental or naturalistic settings as 1D time-dependent filters.

\section{Material and Methods}

\subsection{Data sets}
\label{datasets}
\paragraph{Simulated data}

For the simulated data shown in Figure \ref{fig:Figure_Sim_LG} and \ref{fig:Figure_Sim_LNP}, we used a 2D rank-2 STRF with 30 time frames and 40 pixels in space ($d=1200$) as shown as "ground truth" in both Figures and generated responses with two types of flicker bar stimuli: Gaussian white noise and pink noise. For LNP models, we used a fixed exponential nonlinearity. 

For Figure \ref{fig:Figure_Simulation_lnln}, we used two 20x20 pixels antagonistic center-surround STRF and with Gaussian white noise as stimulus. We used a fixed exponential function for the filter nonlinearity and a softplus function for the output nonlinearity. We controlled the mean firing rate of all simulated spike trains to about 21 Hz by adjusting the intercept of the corresponding models (for Figure \ref{fig:Figure_Sim_LNP}A and C, the intercepts were 3.5 and 2.5, respectively. For Figure \ref{fig:Figure_Simulation_lnln}, the intercept was 0) and also multiplying a constant $R=10$ to $\lambda$. 

For all simulations, we measured the similarity of the ground truth and the model estimators by computing the MSE between the respective normalized STRFs. The STRFs were normalized by element-wise division by the matrix-norm.

For Figure \ref{fig:Figure_Sim_LG}, the similarity was measured for various stimulus length ($n_{\mathrm{samples}}$), which were scaled by different factors ($\mathrm{factor} \in {0.5, 1, 2, 4, 8, 16, 32, 64}$) with respect to the number of coefficients of the STRF. For Figure \ref{fig:Figure_Sim_LNP} and \ref{fig:Figure_Simulation_lnln}, the similarity was measured at various lengths of time ($\mathrm{time} \in {0.5, 1, 2, 4, 8, 16, 32, 64}$ minutes), which is equal to setting $\mathrm{time} = n_{\mathrm{samples}} \times \Delta$, where $\Delta$ is the time bin size and took the value of 0.033 s.

Error bars in Figure \ref{fig:Figure_Sim_LG}-\ref{fig:Figure_Simulation_lnln}
 were obtained by averaging over 10 trials of simulations with different random seeds for the stimulus.
 
For the simulated data shown in Figure \ref{fig:Figure_diagnostics}, we used a 3D STRF with 25 time frames and 25x25 pixels in space ($d=25\times25\times25=15625$), where the spatial profile was a 2D Gaussian field and the temporal profile had a sharp dip close to zero time lag, and generated responses with Gaussian white noise (sample size $n=d$ unless noted otherwise). Additive Gaussian noise was added to the responses ($\sigma=1$ except (c), for which $\sigma=3$).

\paragraph{Experimental data}

We used a previously published data set for Figure \ref{fig:Figure_Expdata}A: extracellularly recorded spikes from tiger salamander retinal ganglion cells \cite{liu_inference_2017}, available to download at \url{https://gin.g-node.org/gollischlab/Liu_etal_2017_RGC_spiketrains_for_STNMF}.

Another two data sets of visual neurons and one data set of a non-visual neuron were used for the supplementary Figure 2 and 3:

\begin{enumerate}
  \item Extracellularly recorded spikes from a tiger salamander retinal ganglion cell \cite{maheswaranathan_inferring_2018}, available to download at \url{https://github.com/baccuslab/inferring-hidden-structure-retinal-circuits};
  \item Extracellular recorded spikes from a V1 complex cell of macaque monkey \cite{rust_spatiotemporal_2005}, available on request from the authors;
  \item Extracellular recorded spikes from a grid cell in rat entorhinal cortex \cite{hafting_microstructure_2005}, available to download at \url{https://www.ntnu.edu/kavli/research/grid-cell-data};
 
\end{enumerate}

The details of the second data set for Figure \ref{fig:Figure_Expdata}B are described below.

\subsection{Experimental procedures for 2-photon calcium imaging}

\paragraph{Animals and tissue preparation}

One wild-type mouse (C57Bl/6J, JAX 000664) housed under a standard 12 hour day/night rhythm was used in this study. Before experiments, the mouse was dark adapted for at least 2 hours. Then, the animal was anaesthetized with isoflurane (Baxter) and killed with cervical dislocation. Eyes were quickly enucleated in carboxygenated (95\% O$_{2}$, 5\% CO$_{2}$) artificial cerebral spinal fluid (ACSF) solution containing (in mM): 125 NaCl, 2.5 KCl, 2 CaCl$_{2}$, 1 MgCl$_{2}$, 1.25 NaH$_{2}$PO$_{4}$, 26 NaHCO$_{3}$, 20 glucose, and 0.5 L-glutamine (pH 7.4). After carefully removed the cornea, sclera and vitreous body, the retina was flattened on an Anodisc (0.2 $\mu$M pore size, GE Healthcare) with the RGC side facing up and electroporated (see below). Then, the retina was transferred to the recording chamber of the microscope. To visualize the blood vessels and the damaged cells, we added 5 $\mu$M sulforhodamine-101 (SR101, Invitrogen) into 2l of ACSF solution. All experimental procedures were carried out under very dim red light. All animal procedures were approved by the governmental review board (Regierungspräsidium Tübingen, Baden-Württemberg, Konrad-Adenauer-Str. 20, 72072 Tübingen, Germany) and performed according to the laws governing animal experimentation issued by the German Government. 

\paragraph{Electroporation}

For two-photon $\mathrm{Ca}^{2+}$ imaging, Oregon-Green BAPTA-1 (OGB-1, hexapotassium salt; Life Technologies) was bulk electroporated \cite{briggman_bulk_2011}. After the retina was flattened on an Anodisc and placed between two 4-mm horizontal plate electrodes (CUY700P4E/L, Nepagene/Xceltis), 10 $\mu$l OGB-1 (5 mM) diluted in ACSF solution was suspended on the upper electrode. Then the upper electrode was lowered onto the retina and applied 10 pulses (~9.2 V, 100 ms pulse width, at 1 Hz) from a pulse generator/wide-band amplifier combination (TGP110 and WA301, Thurlby handar/Farnell). For complete recovery of the electroporated retina, we started recordings 30 min after electroporation. 

\paragraph{Two photon imaging and light stimulation}

A MOM-type two-photon microscope (designed by W. Denk, MPI, Martinsried; purchased from Sutter Instruments/Science Products) as described previously \cite{euler_eyecup_2009, euler_studying_2019} was used for this study. Briefly, the system was equipped with a mode-locked Ti:Sapphire laser (MaiTai-HP DeepSee, Newport Spectra-Physics, Darmstadt, Germany), green and red fluorescence detection channels for OGB-1 (HQ 510/84, AHF, Tübingen, Germany) and SR101 (HQ 630/60, AHF), and a water immersion objective (16x/0.80W, DIC N2, $\infty$/0 WD 3.0, Nikon). For $\mathrm{Ca}^{2+}$ imaging, we tuned the laser to 927 nm, and used a custom-made software (ScanM, by M. Müller, MPI, Martinsried, and T.E.) running under IGOR Pro 6.3 for Windows (Wavemetrics). Time-elapsed $\mathrm{Ca}^{2+}$ signals were recorded with 64$\times$16 pixel image sequences (31.25 Hz). 

Light stimuli were projected through the objective lens \cite{euler_studying_2019}. A digital light processing (DLP) LightCrafter 4500 (Texas Instruments, Dallas, TX) coupled with external unit with light-emitting diodes (LEDs) – "green" (576 nm) and UV (387 nm) that match the spectral sensitivities of mouse M- and S-opsins (for details, see \cite{baden_functional_2016, franke_arbitrary-spectrum_2019}) was used in this study. Both LEDs were intensity-calibrated to range from $\mathrm{0.1 \cdot10}^{3}$ ("black" background) to $\mathrm{20.0 \cdot10}^{3}$ ("white" full field) photoisomerizations $\mathrm{P^{*}s^{-1}}$ per cone. The light stimulus was centered before every experiment, ensuring that its center corresponded to the center of the microscope’s scan field. For all experiments, the tissue was kept at a constant mean stimulator intensity level for $\geq$ 15 s after the laser scanning started and before light stimuli were presented.    

Binary dense noise (20$\times$15 matrix of 30$\times$30 $\mu$M pixels; each pixel displayed an independent, balanced random sequence at 5 Hz for 20 minutes) were generated and presented using the Python-based software package QDSpy (\url{https://github.com/eulerlab/QDSpy}).

\paragraph{Data pre-processing}

$\mathrm{Ca}^{2+}$ imaging data were pre-processed using custom-written scripts in IGOR Pro. Regions-of-interest (ROIs) were manually defined depending on the outline of the soma. For each ROI, $\mathrm{Ca}^{2+}$ traces were extracted using the image analysis toolbox SARFIA for IGOR Pro \cite{dorostkar_computational_2010}. A stimulus time marker embedded in the recorded data served to align the traces relative to the light stimulus at a temporal resolution of 2 ms. All stimulus-aligned traces together with the relative ROI position were exported for further analysis.

\subsection{Previous methods}

\paragraph{Spike-triggered Average (STA)}

The STA was computed as the average of all stimuli that proceeded a spike \cite{aljadeff_analysis_2016}:

    \begin{equation} 
        w_{STA} = \frac{1}{n} X^Ty \label{eq_sta}
    \end{equation}

where $X$ is the stimulus design matrix, $y$ the response, $n$ can be the number of spikes or the number of samples (rows) in $X$ and $y$ for non-spike data. 

\paragraph{Whitened spike-triggered average (wSTA)} 

The maximum-likelihood estimator under a Gaussian noise model, or whitened-STA (wSTA), was computed by multiplying the STA with the inverse of the stimulus auto-correlation matrix:

    \begin{equation} 
        w_{wSTA} = (X^TX)^{-1}X^Ty \label{eq_wsta}
    \end{equation}
    
In our code base, we did not compute the matrix inverse explicitly but computed the least squared solution instead, which is common practice.
    
\paragraph{Maximum-a-posteriori estimator (MAP)} 

Similar, the MAP under a Gaussian noise model was computed by adding the inverse of the prior covariance to the stimulus auto-correlation matrix:

    \begin{equation} 
        w_{MAP} = \left(X^TX + C^{-1} \right)^{-1}X^Ty \label{eq_map}
    \end{equation}

where $C$ is a prior covariance matrix of choice, e.g. corresponding to a smoothness or sparseness assumption \cite{sahani_evidence_2002, park_receptive_2011}. Its hyperparameters can be optimized by minimizing the negative log-evidence\cite{sahani_evidence_2002, park_receptive_2011}:

    \begin{equation} 
        \mathrm{Cost}_{MAP} = \frac{n}{2}\mathrm{log}|2\pi\sigma^2| 
        + \frac{1}{n}\mathrm{log}|C\Lambda^{-1}| - \frac{1}{n}\mu^T\Lambda\mu
        + \frac{1}{2\sigma^2}Y^TY
    \end{equation}

where $\sigma$ is variance of the additive Gaussian noise, $\mu$ and $\Lambda$ are posterior mean and covariance, respectively:

    \begin{equation} 
        \mu = \frac{1}{\sigma^2}\Lambda X^Ty
    \end{equation}

and

    \begin{equation} 
        \Lambda = \left(\frac{1}{\sigma^2}X^TX + C^{-1} \right)^{-1}
    \end{equation}    
    
\paragraph{Smoothness prior covariance matrix for ASD} The 1D ASD prior covariance matrix is formulated as following \cite{sahani_evidence_2002}:

    \begin{equation} 
        C_{ij} = \exp\left(-\rho - \frac{\Delta_{ij}}{2 \delta ^2} \right) 
    \end{equation}

where $\Delta_{ij}$ is the squared distance between STRF coefficients $w_i$ and $w_j$, $\rho$ controls the scale and $\delta$ controls the smoothness.

\paragraph{Locality prior covariance matrix for ALD} The 1D ALD prior covariance matrix is formulated as following \cite{park_receptive_2011}:

    \begin{equation} 
        C = C_{s}^{\frac{1}{2}}(B^T C_{f} B)C_{s}^{\frac{1}{2}}
    \end{equation}

where $B$ is an orthogonal basis matrix for the 1D discrete Fourier transform, $C_s$ and $C_f$ are the diagonal ALD prior covariance in space and frequency, respectively, which can be constructed as:

    \begin{equation} 
        C_{s, ii} = \exp\left(- \frac{1}{2\tau_s^2}(\chi_{i} - \nu_s)^2)\right)
    \end{equation}
    
where $\tau_s$ is scalar values control the shape and extent of the local region, $\chi$ is the STRF coefficient coordinates in space-time and $\nu_s$ is center coordinate of the STRF. And

    \begin{equation} 
        C_{f, ii} = \exp\left(- \frac{1}{2} (|\tau_f \omega_i| - \nu_f)^2)\right)
    \end{equation}
    
where $\tau_s$ is scalar values control the scale of smoothness in frequency, $\omega$ is of STRF coefficient coordinates in frequency and $\nu_f$ is center coordinate of the STRF in Fourier domain.

High-dimensional prior covariance matrix can be approximated by taking the Kronnecker product ($\otimes$) of the covariance matrix in each dimension.
    
\subsection{Estimating STRFs with Natural Cubic Regression Splines}
\label{method-cubic-spline}
To generate a natural cubic spline matrix ($S$), we used the implementation from \textsc{Patsy} (0.5.1), a Python equivalent of the original implementation in R package \textsc{mcgv} \cite{wood_generalized_2017}. The only hyperparameter needed to generate a 1D spline basis matrix was the number of basis functions, or the degrees of freedom (df), as we set all knots to be equally spaced. 

\begin{equation} 
    S = a^{-}_{j}(x) + a^{+}_{j}(x) + c^{-}_{j}(x)F_j + c^{+}_{j}(x)F_{j+1} \; \mathrm{if} \; x_j < x \leq x < x_{j+1}
\end{equation}

where $S$ is the spline basis matrix, $x$ is knot locations, $F^{-1} = B^{-1}D$, $F = (0, F^{-1}, 0)$ and $a^{-}_{j}$, $a^{+}_{j}$, $c^{-}_{j}$, $c^{+}_{j}$, $B$, $D$ are defined in Table \ref{tab:cr_spline}. 

    \begin{table}
        \centering
        \begin{tabular}{llll}
            \\
            $h_j = x_{j+1} -x_j$ & \\
            $a^{-}_{j}(x)=(x_{j+1}-x)/h_j$ & $c^{-}_{j}(x)=[(x_{j+1}-x)^3/h_j - h_j(x_{j+1}-x)]/6$ & \\
            $a^{+}_{j}(x)=(x-x_{j})/h_j$  & $c^{+}_{j}(x)=[(x-x_{j})^3/h_j - h_j(x-x_{j})]/6$ & \\
            $D_{i,i}=1/h_i$ & $D_{i, i+1}=-1/h_i - 1/h_{i+1}$ & $D_{i, i+2}=1/h_{i+1}$\\
            $B_{i,i}=(h_i+h_{i+1})/3$ & & i=1...k-2 \\ 
            $B_{i, i+1}=h_{i+1}/6$ & $B_{i+1, i} = h_{i+1} /6$ & i=1...k-3
        \end{tabular}
        \caption{Basis functions for natural cubic spline.}
        \label{tab:cr_spline}
    \end{table}

An illustration of 1D and 2D natural cubic splines can be seen in Figure  \ref{fig:Figure_basis_example}B. A spline-based STRF under a Gaussian noise model can be fitted in the similar manner as the wSTA:

    \begin{equation} 
        b_{SPL, wSTA} = (S^TX^TXS)^{-1}S^TX^Ty \label{eq_spl_mle}
    \end{equation}

where $b$ is the spline coefficients, and the corresponding STRF is:

    \begin{equation} 
        w_{SPL, wSTA} = Sb_{SPL, wSTA}=S(S^TX^TXS)^{-1}S^TX^Ty
    \end{equation}

The spline matrix can be extended to high dimensions, which is also called tensor product smooth \cite{wood_generalized_2017}, by simply taking the Kronnecker product ($\otimes$) of the basis functions in each dimension:

    \begin{equation} 
        S = S_t \otimes S_x \otimes S_y
    \end{equation}

\subsection{Spline-based GLMs with gradient descent} 
\label{method-spline-glm}
The spline coefficients can also be fitted by gradient descent with respect to the cost function defined in each model with L1 regularization. 

For the LG model, the cost function is the sum square error:

    \begin{equation} 
        \mathrm{Cost}_{LG} = \frac{1}{n}\sum( y - XSb)^2  + \alpha\Vert b \Vert_{1}
    \end{equation}

where $\alpha$ is the weight of L1 regularization. 

For LNP and LNLN model, the cost function is the negative log-likelihood:

    \begin{equation} 
        \begin{split}
            \lambda = f(XSb), \\
            \mathrm{Cost}_{LNP / LNLN} = -ylog(\lambda) + \sum \Delta\lambda + \alpha\Vert b \Vert_{1}
        \end{split}
    \end{equation}

where $\lambda$ is the conditional intensity or the instantaneous firing rate, $\Delta$ is the size of time bin, and $f(.)$ is a nonlinearity applied to the filter output, which can be fixed as a softplus or exponential nonlinearity.
 
We used JAX \cite{bradbury_jax_2018} for automatic differentiation, and Adam optimizer \cite{kingma_adam_2017}for parameters optimization.

\subsection{Spline-based spike-triggered clustering methods}
\label{method-spike-triggered-clustering}
Both k-means clustering and semi-NMF can be formulated as a matrix factorization problem:

    \begin{equation} 
        \mathrm{min} \Vert V_{\pm} - W_{\pm}H_{+}^{T} \Vert^{2}
    \end{equation}

where $V$ is the spike-triggered ensemble, $W$ is clustering centroids matrix (or subunit matrix), and $H$ is the clustering label (or subunit weight matrix). A spline basis can be easily incorporated into k-Means clustering by assuming that each centroid can be approximated by the same spline basis matrix: $W_{SPL, centroid} = SB$, where $B$ is the spline coefficient matrix for each subunit, and can be computed as the least-squares solution to the linear matrix equation $B=S^{\dagger}W_{centroid}$, where $\dagger$ is pseudo-inverse. For semi-NMF, splines can also be combined with the update rules \cite{ding_convex_2010}:

    \begin{subequations}
      \begin{equation}
          W = SS^{\dagger}VH(H^TH)^{-1} 
      \end{equation}
      \begin{equation}
            H = H * \sqrt{\frac{(V^TW)^{+}+[H(W^TW)^{-}]}{(V^TW)^{-}+[H(W^TW)^{+}]}}  
      \end{equation}    
    \end{subequations}

where $*$ is point-wise multiplication, $A^{+}=(\mid A\mid + A)/2$ and $A^{-}=(\mid A\mid - A)/2$.

\subsection{Confidence interval and statistical significance of the estimated STRFs}
\label{method-ci}
The asymptotic posterior probability of the estimated STRF coefficients can be computed analytically \cite{wood_generalized_2017}: $b|y, X \sim N(b, V_{b})$, where $b$ is the fitted STRF coefficients, $V_{b}$ is the posterior covariance. For LG model:

    \begin{equation} 
        V_{b} = (S^T X^T XS)^{-1}\sigma^2
    \end{equation}

where $\sigma^2$ is the residual sum of squares of the data and predicted response from the training set. And for LNP model:

    \begin{equation} 
        V_{b} = (S^T X^TW XS)^{-1}
    \end{equation}

where $W$ is a diagonal matrix with entries $w_{ii} = \frac{1}{f(XSb)^{2}}$, and $f(.)$ is the nonlinearity of the model. Thus, the standard error of $b$ is the square root of the diagonal of $V_{b}$, and the corresponding confidence interval (CI) of the estimated STRF is as following:

    \begin{equation} 
        b - 1.96 \times b_{se} \leq CI \leq b + 1.96 \times b_{se} 
    \end{equation}

Given the computed confidence interval, Wald Chi-Squared Test can be used to determine whether a STRF is statistically significant (against the null hypothesis that all STRF coefficients are zero), with the Wald statistic $T = bV_{b}^{-1}b^{T}$.

\subsection{Permutation Test}
\label{method-permutation}
To evaluate the predictive performance of the estimated STRF, we permuted the validation test stimulus in time and made prediction over 100 repetitions. A one-tailed one-sample Student's T-test was used to evaluate if the model prediction was better than chance level. 

\subsection{Visualization of 3D STRF}
\label{method-visualization}

For visualization of 3D STRF in Figure \ref{fig:Figure_diagnostics} and \ref{fig:Figure_Expdata}, we first performed singular value decomposition (SVD) to separate the spatial and temporal components of the estimated 3D STRF. The coordinates of the extremum pixel in the spatial component were selected, and the temporal filter of the original STRF is shown for those coordinates. The confidence interval of the same extremum point was plotted as a gray shaded region on the temporal filter. For the spatial filter, we show the frame of the original STRF at the extremum point of the temporal filter (indicated by a vertical dotted line). The maximum value of the color map was set to the maximum absolute value of the STRF.

\subsection{Comparison of Computation Time}
\label{method-computation-time}
For comparison of computation time in Figure \ref{fig:time_comparison}, we used simulated 2D RFs of different sizes (15x15, 20x20 and 30x30 pixels, respectively) and white noise stimuli. Simulations were carried out at different lengths of time in the same manner as Figure \ref{fig:Figure_Sim_LNP} and \ref{fig:Figure_Simulation_lnln}. The number of basis functions for SPL wSTA were fixed as 5 and 10 along each dimension in all three cases. Measurements of computation time were averaged over 10 repetitions and performed on a 2019 16-inch MacBook Pro with a 2.3 GHz 8-core Intel Core i9 and 16 GB of RAM.

\subsection{Data availability}

All data used are available in GitHub (\url{https://github.com/huangziwei/data_RFEst}). 

\subsection{Code availability}

All methods (spline-based GLMs, evidence optimization and spike-clustering methods) were implemented in Python3 and available in GitHub (\url{https://github.com/berenslab/RFEst}). And Jupyter notebooks for all figures are available also in GitHub (\url{https://github.com/huangziwei/notebooks_RFEst})

\section*{Acknowledgements}
We thank the authors of \cite{maheswaranathan_inferring_2018, liu_inference_2017, rust_spatiotemporal_2005} and \cite{hafting_microstructure_2005} for making their data available. Research was funded by the Deutsche Forschungsgemeinschaft through a Heisenberg Professorship (BE5601/4-1 and BE5601/8-1, PB), the Collaborative Research Center 1233 ``Robust Vision'' (reference number 276693517) as well as individual research grants (EU 42/10-1, BE5601/6-1), the German Ministry of Education and Research through the Bernstein Award (01GQ1601, PB). PB is a member of the Excellence Cluster 2064 ``Machine Learning --- New Perspectives for Science'' (reference number 390727645) and the Tübingen AI Center (FKZ: 01IS18039A).


\printendnotes
\bibliography{references_submit}

\end{document}


\maketitle

\section{The number of spline basis functions vs. the amount of data used for inference.}

We investigated how different amounts of training data affects the optimal number of basis (df) estimated from cross-validation. To this end, we used the complete recording of the example cell from Figure 9A. The last 2 minutes of the recording were used as validation set. The rest of the recording was then divided into different lengths (1, 2, 4, 8, 16, 32 and 64 minutes). We used the same procedure of grid search from Figure 8 for all recording lengths. We used a smaller grid for this analysis: from 6 to 11 for temporal dimension and from 8 to 11 for spatial dimension, as we observed from Figure 8 that the optimal values would most likely fall into these ranges. 

For most lengths, the optimal df was [9, 10, 10]. When only 1 minutes of the data was used, the optimal number of basis for both dimensions were relatively smaller than others (df=[8,8,8]). On the other hand, when about 1 hour of the data was used, the df for spatial dimension increased one, while the df for temporal dimension was still the same. Overall, the effect of the amount of data to the optimal df was not very strong.

    \begin{figure}[htp]
        \centering
        \includegraphics[width=\textwidth]{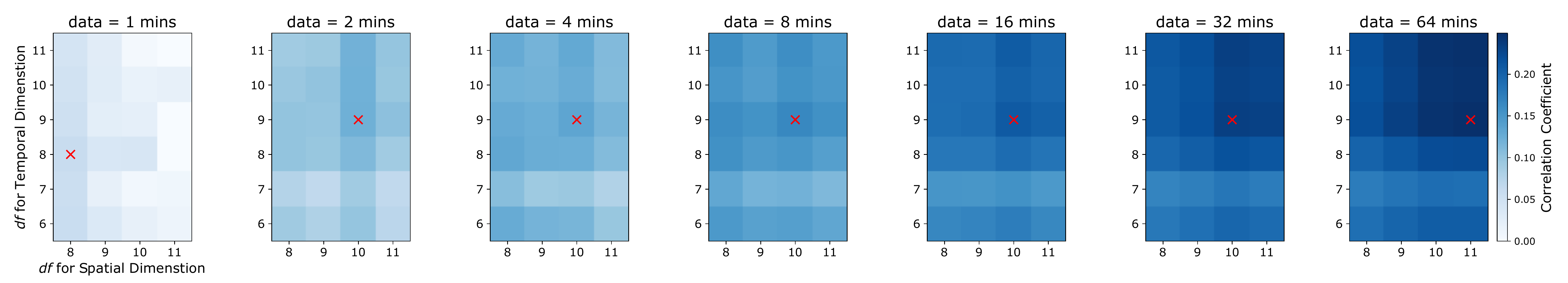}
    
        \caption{The number of spline basis functions vs. the amount of data used for inference. The predictive performance of the MLE on the validation set was plotted as a 2D heat map, and the optimal dfs were indicated by the red crosses.}
        \label{fig:SupFigure_CV}
    \end{figure}

\newpage
\section{More experimental data examples}

We applied spline-based methods to two more previously published data sets of visual neurons and one data set of non-visual neuron:

\begin{enumerate}
  \item Extracellularly recorded spikes from a tiger salamander retinal ganglion cell \cite{maheswaranathan_inferring_2018}, available to download at \url{https://github.com/baccuslab/inferring-hidden-structure-retinal-circuits};
  \item Extracellular recorded spikes from a V1 complex cell of macaque monkey \cite{rust_spatiotemporal_2005}, available on request from the authors;
  \item Extracellular recorded spikes from a grid cell in rat entorhinal cortex, available to download at \url{https://www.ntnu.edu/kavli/research/grid-cell-data}; \cite{hafting_microstructure_2005};
 
\end{enumerate}

Data preparation was the same as Figure 9 in the main text: the first 10 minutes of the whole recording were used for fitting the models, another 2 minutes were used for model validation and another 4 sets of 2 minutes for testing. Number of subunits were also set to 4 through out for the first two data sets.

\begin{figure}[b]
        \centering
        \includegraphics[width=\textwidth]{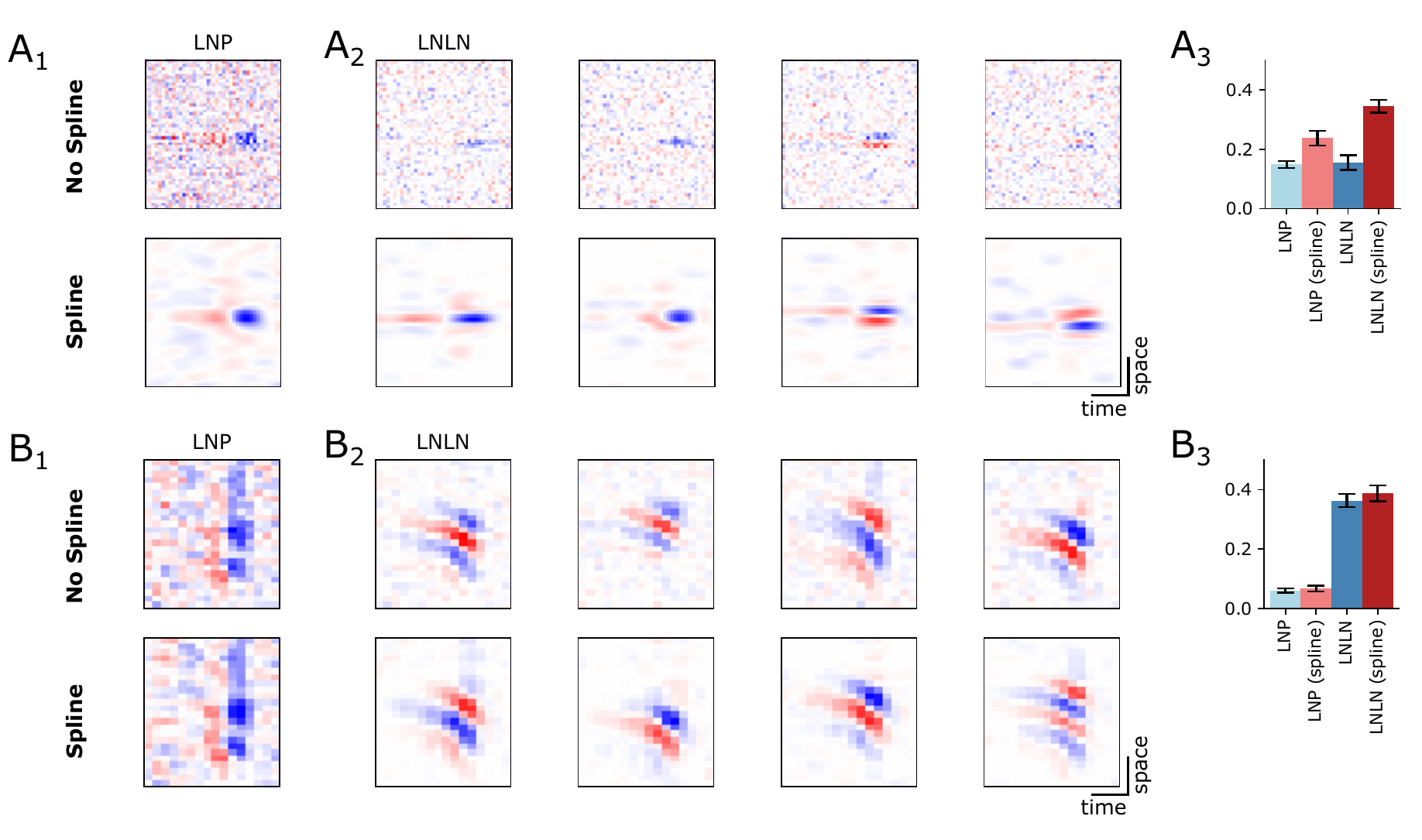}
    
        \caption{Application of the spline-based LNP and LNLN to two data sets of visual neurons. \textbf{A} Extracellularly recorded spikes from a salamander RGC from\cite{maheswaranathan_inferring_2018} using LNP and LNLN models, as well as their predictive performance. \textbf{B} As in A. but for a complex cell in visual cortex from \cite{rust_spatiotemporal_2005}. }
        \label{fig:SupFigure_Expdata}
    \end{figure}

For the example neuron from data set (1), the spline-based LNP and LNLN outperformed their non-spline versions (Supplementary Figure  \ref{fig:SupFigure_Expdata}A), similar to the examples shown in Figure 9 in the main text. In contrast, the improvement of spline-based methods for the example neuron from data set (2) were not as obvious (Supplementary Figure  \ref{fig:SupFigure_Expdata}B). One possible reason was that the signal-to-noise ratio of this data set was much higher than other data sets, as the total number of spikes in the training set were 10 times more than for data set (1). Therefore, the non-spline version was capable of fitting a reasonable STRF even without the spline basis. The second reason might be that the number of subunits were under-estimated in our example (the total number of subunits reported was more than 7 in the original paper).

\begin{figure}[htp]
        \centering
        \includegraphics[width=\textwidth]{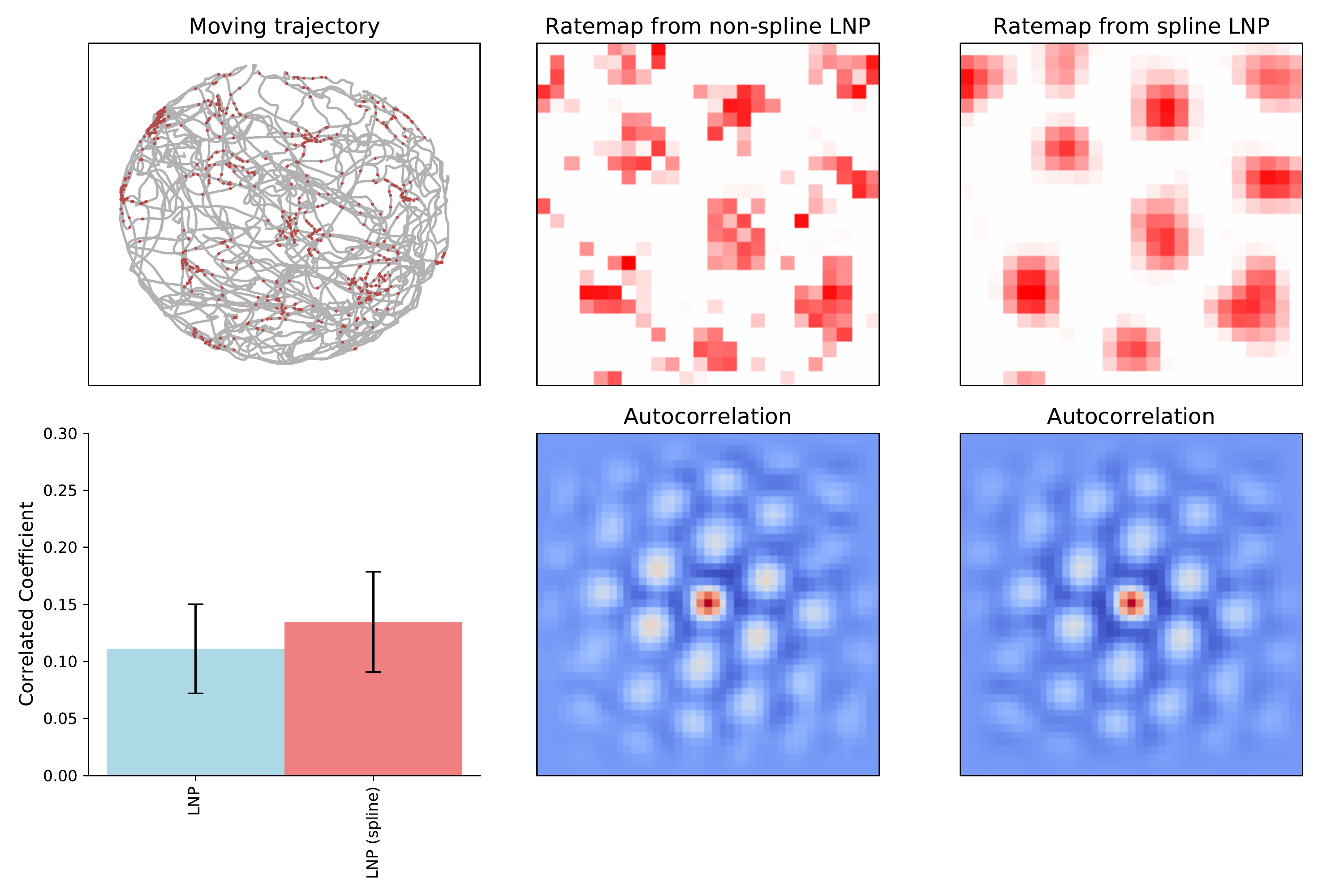}
    
        \caption{Application of a spline-based LNP to a grid cell from  \cite{hafting_microstructure_2005}. Top, left: Trajectory of the animal (grey) with superimposed spikes (red). Middle: Grid field estimated using a non-spline LNP model. Right: Grid field estimated using the spline LNP model. Bottom, left: Predictive performance of the two models. Middle and right: Autocorrelation function computed from the rate maps above.}
        \label{fig:SupFigure_Gridcell}
    \end{figure}

Finally, we applied the spline LNP to grid cells from entorhinal cortex (Supplementary Figure \ref{fig:SupFigure_Gridcell}). A LNP model might be too simplistic to account for the firing of a grid cell (as reflected in the mediocre model prediction performance), but such model can still be fitted with the moving trajectory of the animal as input (the coordinates of the trajectory were binned into a 24 by 24 grid). One obvious advantage of using a spline LNP for retrieving the grid field is that the estimated field is automatically smoothed. Also, locations the animal rarely visited could become visible, whereas they would most likely be not visible in a grid field estimated from non-spline LNP or STA.

\bibliographystyle{unsrt}
\bibliography{references}